\documentclass{article} %
\usepackage{iclr2027_conference,times}

\usepackage{amsmath,amsfonts,bm}

\def\eqref#1{equation~\ref{#1}}

\def\1{\bm{1}}

\DeclareMathAlphabet{\mathsfit}{\encodingdefault}{\sfdefault}{m}{sl}
\SetMathAlphabet{\mathsfit}{bold}{\encodingdefault}{\sfdefault}{bx}{n}

\usepackage{hyperref}
\usepackage{url}
\usepackage{graphicx}
\usepackage{amsmath}
\usepackage{subcaption}
\usepackage[dvipsnames]{xcolor}
\usepackage{soul}
\usepackage{algorithm}
\usepackage{algpseudocode}
\usepackage{amssymb}
\usepackage{wrapfig}
\usepackage{booktabs}

\title{Local Support Learning}

\author{\textbf{Assaf Ben-Kish}$^{\text{\textbf{1}} \dagger}$, \textbf{Akarsh Kumar\textsuperscript{2}}, \textbf{James Glass\textsuperscript{2}}, \textbf{Raja Giryes\textsuperscript{1}} \\
\hspace{0.05em} \textsuperscript{1}Tel Aviv University,  \textsuperscript{2}MIT CSAIL
}

\iclrfinalcopy %
\begin{document}

\maketitle

{\renewcommand{\thefootnote}{\fnsymbol{footnote}}\footnotetext[2]{Work done while visiting the Spoken Language Systems group at MIT CSAIL}}

\vspace{-10pt}

\begin{abstract}

We explore catastrophic forgetting in the context of large pre-trained models. By considering forgetting as a geometric problem in the input space of each weight matrix, we uncover a natural retention objective under which updates produced by gradient-based optimizers are suboptimal. Following this observation, we propose Local Support Learning (LSL), a general-purpose framework that augments gradient-based training for retention of prior capabilities without access to prior data.
During a new learning phase, LSL pairs two components with distinct roles: a standard weight adapter, trained as usual to minimize the loss, and a gating function that enables the adapter only on input activations from its own training distribution, making the update \emph{local} to that distribution.
The key challenge is that this gate must route data from all learning phases while training only on data from the current one. We address this with a gate based on a Gaussian Mixture Model (GMM), whose likelihood decays rapidly away from its training data, giving it a natural tendency to stay closed on data from prior phases. We show that this post-training approach can resolve forgetting in LLMs of up to 7 billion parameters, retaining both pretrained and finetuned capabilities across multiple training phases, while being efficient in memory and compute, robust to hyperparameter choice, and showing scaling potential. \newline
Website and code: \url{https://assafbk.github.io/lsl}

\end{abstract}

\begin{figure}[h]
  \centering
  \includegraphics[width=0.9\textwidth, trim=0 95 0 95, clip]{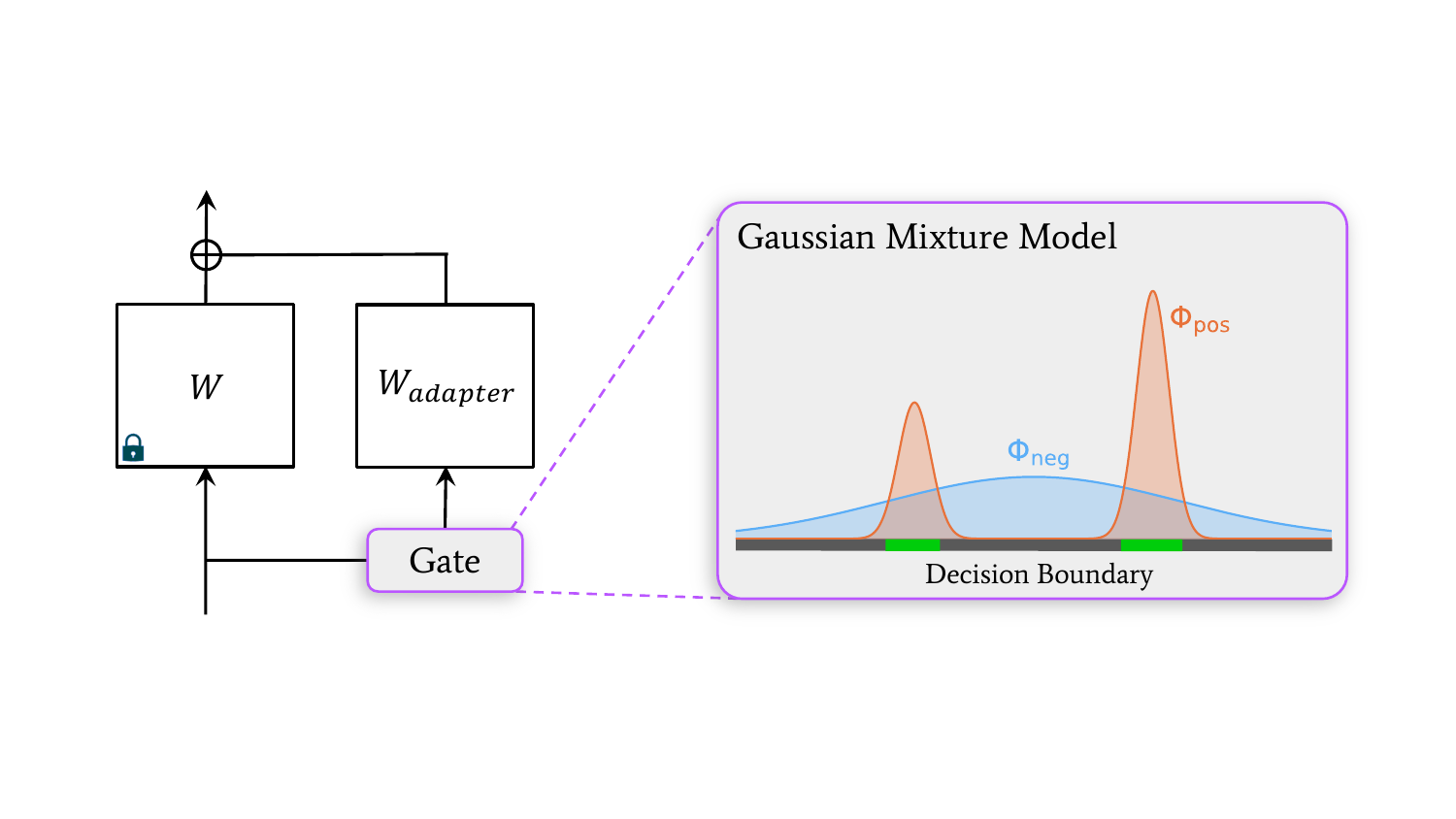}
  \caption{\textbf{Avoiding catastrophic forgetting with Local Support Learning.} (\textbf{Left}) We augment gradient-based training with a mechanism that retains prior capabilities without access to prior data. The key component is a gating function that enables the weight adapter only on input activations from its training distribution. Interestingly, conventional MLP classifiers are not well suited for this task. Instead, we propose a GMM-based gate that tends to stay closed on prior data it has never encountered.
  (\textbf{Right}) 1D illustration of the GMM-based gate. $\Phi_{pos}$ (orange) 
  captures the distribution of the finetuning data, while the wider $\Phi_{neg}$ (blue) dominates elsewhere - ensuring that data outside the current distribution is routed only to the pretrained weights. 
}
  \label{fig:method}

  \vspace{-6pt}
\end{figure}

\begin{figure}[h]
  \centering
  \begin{subfigure}{\textwidth}
    \centering

    \includegraphics[width=\textwidth, trim=5 135 10 105, clip]{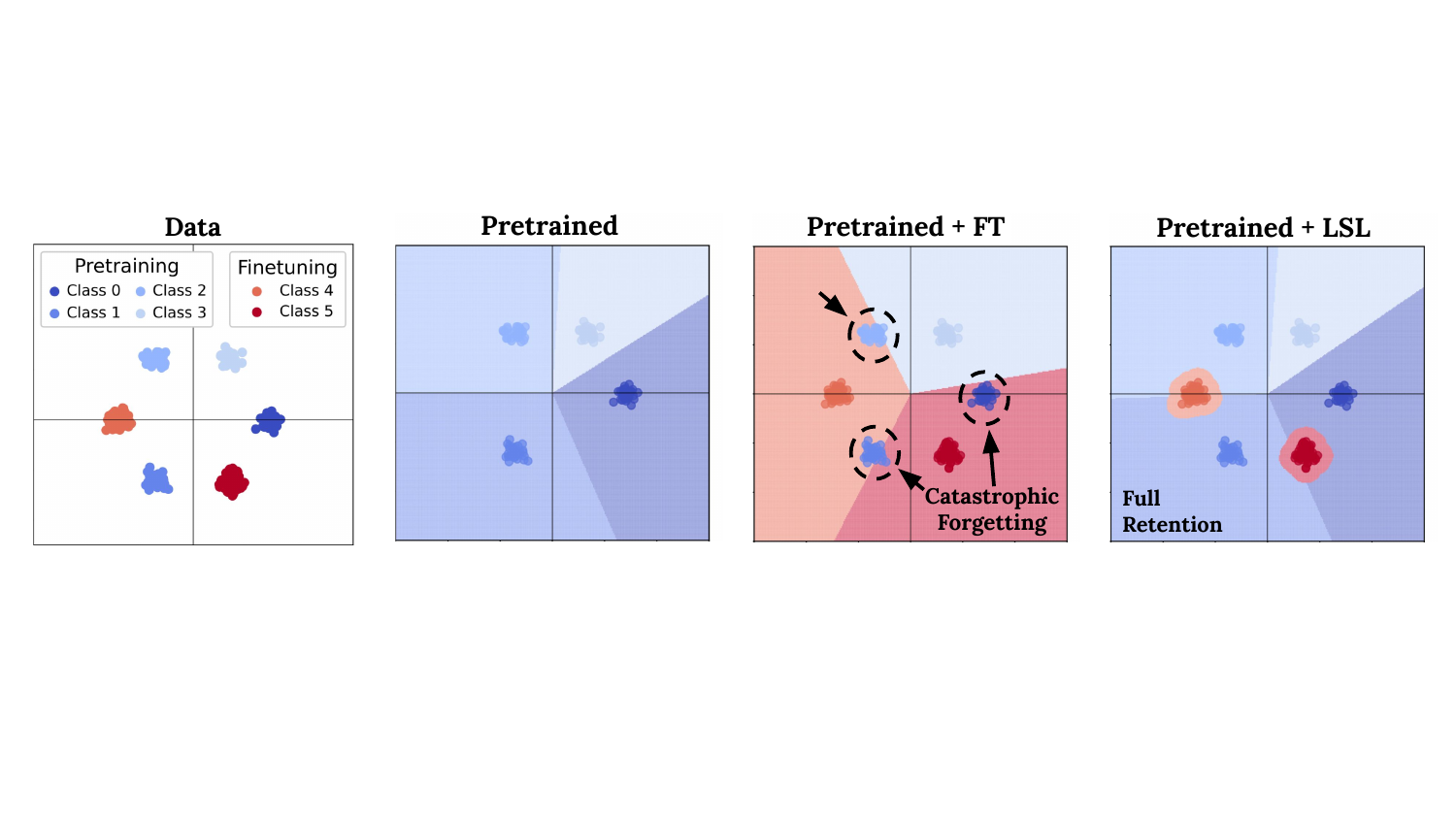}
    
    \label{fig:toy_problem_data}
  \end{subfigure}
  \caption{\textbf{Catastrophic forgetting - toy example}. (\textbf{Data}) We train over six classes in two phases: ``Pretraining" (points in blue), and ``Finetuning" (points in red). (\textbf{Pretrained}) Classifier trained on the pretraining data. The background colors are the learned decision boundaries - depicting a perfect score on the observed distribution. (\textbf{Pretrained + FT}) We continually train the model on two other classes. The model performs well on data from the second phase (red), yet the boundaries learned during the first phase are heavily deformed - a clear case of forgetting. (\textbf{Pretrained + LSL}) Same experiment, this time with LSL. We restrict weight updates to the region of the input space that produced them, leaving model behavior unchanged elsewhere - thus achieving a perfect score.}
  
  \vspace{-14pt}
  
  \label{fig:both}
\end{figure}

\section{Introduction}

The modern deep learning paradigm first pretrains a neural network on a large dataset and then adapts it to desired tasks and behaviors through additional finetuning phases~\citep{ouyang2022traininglanguagemodelsfollow, wei2022finetunedlanguagemodelszeroshot, rozière2024codellamaopenfoundation}. However, neural networks tend to forget previously learned capabilities when adapting to new data - a phenomenon known as catastrophic forgetting~\citep{MCCLOSKEY1989109}. 
Forgetting hinders multi-phase learning, as capabilities acquired during pretraining and previous finetuning phases often get overwritten~\citep{biderman2024loralearnsforgets}.

Catastrophic forgetting is a long-standing problem, studied well before pretraining became popular.
Among earlier mitigation approaches, the most successful augment the model with buffers of previous data samples~\citep{chaudhry2019efficientlifelonglearningagem}, apply data-dependent regularization~\citep{ewc2017}, or impose optimization constraints~\citep{ogd2020}. 
However, these auxiliary mechanisms scale poorly, making them impractical in the modern paradigm, especially for retaining capabilities from the large-scale pretraining phase. Instead, mitigation approaches now aim to minimize change in pre-trained networks~\citep{xiong2025oploraorthogonalprojectionlora,nayak2025sculptingsubspacesconstrainedfinetuning, wang2023orthogonalsubspacelearninglanguage, liang2024inflorainterferencefreelowrankadaptation}, yet the exact definition of change, and how to prevent it, remain open problems.

By considering change at the level of a single weight matrix, we can analyze the effect of an update on the full input-output mapping of the matrix. %
In the context of large-scale pretraining, this perspective yields a natural retention objective: updates should affect only the input region required for learning, i.e., the support of the current phase's distribution. We show that updates produced by gradient-based optimizers often act outside this support: any prior input not orthogonal to the update has its mapping altered, which may lead to forgetting. Existing works that enforce orthogonality to all prior inputs work in small-scale settings~\citep{saha2021gradientprojectionmemorycontinual}, yet become infeasible at pretraining scale.
Without assuming access to pretraining data, recent approaches either protect only previous finetuning phases, keeping updates orthogonal to them~\citep{wang2023orthogonalsubspacelearninglanguage,liang2024inflorainterferencefreelowrankadaptation}, or approximate the orthogonal complement of the pretraining data from the weights~\citep{xiong2025oploraorthogonalprojectionlora,nayak2025sculptingsubspacesconstrainedfinetuning} - an approximation we find insufficient (App.~\ref{app:op_lora_performance}), calling for a new approach.

Motivated by this, we propose Local Support Learning (LSL), a general-purpose framework that augments standard gradient-based training to retain prior capabilities without access to prior data. During a new learning phase, LSL pairs two components with distinct roles: a standard weight adapter~\citep{hu2021loralowrankadaptationlarge}, trained as usual to minimize the loss, and a gating function that estimates the support of the current data in the adapter's input space. At inference, the gate applies the adapter only to tokens whose activations fall within that support. It is a difficult challenge given that the gate cannot train on data from prior phases.
We address this with a gate based on Gaussian Mixture Models (GMMs). The key idea is that a Gaussian's density decays exponentially with distance from the training data, providing the gate an inductive bias to stay closed on inputs it has never seen. %

To demonstrate the effectiveness of our approach, we finetune LLMs of up to 7 billion parameters on diverse tasks and show that they are able to learn at full capacity without degrading capabilities acquired during pretraining including math, coding, and instruction following. This holds even across multiple finetuning phases, where LSL retains both pretrained and previously finetuned capabilities. In addition, retention is largely unaffected by the choice of hyperparameters. This is in contrast to existing methods, where the learning rate, adapter rank, and batch size induce a direct trade-off between learning and forgetting. On the practical side, LSL exhibits desirable scaling trends and is highly efficient in both memory and compute. It requires no more memory than a LoRA adapter while offering fast inference, competitive training time, and straightforward parallelization.

\textbf{Our main contributions} are four-fold: 
(i) We analyze forgetting in modern large-scale settings and uncover a natural retention objective that existing methods do not satisfy, providing a possible explanation for their limited ability to mitigate forgetting.
(ii) We propose Local Support Learning, a general-purpose framework that avoids forgetting by restricting weight updates to the region of the input space that produced them.
(iii) We motivate our approach by providing a theoretical connection between optimal gates and density estimators.
(iv) We evaluate LSL at scales of up to 7 billion parameters, demonstrating its effectiveness, robustness, and efficiency.

\section{Related Work}
\label{sec:rw}

\noindent \textbf{Modern Learning Paradigms.} Large-scale pretraining has proven transformative across modalities, enabling in-context learning in language~\citep{brown2020languagemodelsfewshotlearners}, robust representations in vision~\citep{dosovitskiy2021imageworth16x16words, radford2021learningtransferablevisualmodels}, and general-purpose features in speech~\citep{baevski2020wav2vec20frameworkselfsupervised, radford2022robustspeechrecognitionlargescale}. Development typically proceeds in phases: large-scale pretraining, followed by post-training that adapts the model to desired tasks and  behaviors: instruction following~\citep{wei2022finetunedlanguagemodelszeroshot, chung2022scalinginstructionfinetunedlanguagemodels}, alignment via RLHF~\citep{ouyang2022traininglanguagemodelsfollow}, domain specialization~\citep{rozière2024codellamaopenfoundation, singhal2023expertlevelmedicalquestionanswering}, and multilingual adaptation~\citep{ustun-etal-2024-aya}. While essential, post-training risks overwriting the very capabilities that pretraining established at enormous cost~\citep{biderman2024loralearnsforgets}. Our work directly addresses this tension.

\noindent \textbf{Catastrophic Forgetting.} The degradation of previously learned capabilities when learning from new data has been studied as a sequential task learning problem since its identification~\citep{MCCLOSKEY1989109}. Earlier approaches either augment the model with buffers of previous data samples~\citep{chaudhry2019efficientlifelonglearningagem,  gem2017, chaudhry2019tinyepisodicmemoriescontinual}, apply data-dependent regularization techniques~\citep{ewc2017, aljundi2018memoryawaresynapseslearning, zenke2017continuallearningsynapticintelligence}, or optimize within subspaces orthogonal to gradients or data from previous tasks~\citep{ogd2020, saha2021gradientprojectionmemorycontinual}. %
However, modern practice introduces a fundamental asymmetry: the first `task' is pretraining on trillions of tokens, followed by several smaller finetuning phases.
This renders earlier approaches impractical: a replay buffer representative of pretraining is infeasible to curate, and the data is often unavailable. Still, these works pinpoint key mechanisms of forgetting; in particular, the analysis of gradient interference~\citep{zeng2019owm, wang2021trainingnetworksnullspace, saha2021gradientprojectionmemorycontinual} shows how a weight update to a single matrix affects all possible inputs to that matrix. In Sec.~\ref{sec:prob_inv} we revisit this analysis under modern constraints and show that updates produced by gradient-based methods can hurt retention of prior capabilities when applied to all inputs.

\noindent \textbf{Catastrophic Forgetting in Large Pretrained Models.} More modern methods try to minimize change in pretrained models. A common approach restricts change in weight space:
Low-Rank Adapters (LoRA)~\citep{hu2021loralowrankadaptationlarge} accumulate updates in a small subspace, forgetting less than full fine-tuning but still forgetting, with the price of reduced learning capacity~\citep{biderman2024loralearnsforgets}.
Other works constrain updates to subspaces orthogonal to the top-$k$ singular vectors of the pretrained weights~\citep{xiong2025oploraorthogonalprojectionlora, nayak2025sculptingsubspacesconstrainedfinetuning}, though the choice of $k$ is heuristic and the remaining vectors may still carry important information. A related line restricts different adapters to orthogonal subspaces~\citep{wang2023orthogonalsubspacelearninglanguage, liang2024inflorainterferencefreelowrankadaptation}, mitigating forgetting between finetuning phases but not of pretraining capabilities.
Learning without Forgetting (LwF)~\citep{li2017learningforgetting} takes a different approach, defining change as a shift in the model's functional behavior. This is enforced by penalizing deviation from the base model's outputs on the finetuning data via knowledge distillation~\citep{hinton2015distillingknowledgeneuralnetwork}. What unites these methods is that they apply gradient updates to all inputs, which, as shown in Sec.~\ref{sec:prob_inv}, causes forgetting. We avoid this by augmenting the gradient update with a gate that restricts its effect to activations from its training distribution. Lastly, due to the abundance of relevant work, we refer the reader to an extended discussion in App.~\ref{appendix:extended_rw}.

\section{Formulation and Problem Investigation}
\label{sec:prob_inv}

\noindent \textbf{Formulation.} We formalize catastrophic forgetting over a sequence of $P$ learning phases. The evaluation is the average performance on all tasks from all phases. Within each phase, the model has IID access to the corresponding dataset; across phases, learning must proceed in a streaming manner, and the memory retained per phase must be bounded. This memory constraint is essential: without it, continual learning admits the trivial solution of storing all previously seen data in a replay buffer and repeatedly training on it, which is impractical at pretraining scale and circumvents the core challenge of learning continually with the model alone. In practice, a small per-phase memory footprint is what would allow scaling to many phases.

\noindent \textbf{Reproducing Catastrophic Forgetting in a Controlled Setup.} To motivate our approach, we study a toy classification problem with six classes over a two dimensional input space $x \in \mathbb{R}^2$. This analysis builds on the per-matrix view of interference~\citep{saha2021gradientprojectionmemorycontinual}, described in Sec.~\ref{sec:rw}. As shown in Fig.~\ref{fig:both} (``Data"), our setup includes two learning phases - ``pretraining" and ``finetuning"\footnote{Interference can arise between any two learning phases; without loss of generality, we adopt the terminology of pretraining and finetuning since retaining pretraining capabilities is the hardest case.}.
In the pretraining phase the model was trained over IID samples from four classes - colored in different shades of blue. In the finetuning phase the model trains on samples from two additional classes - colored in different shades of red. We denote both distributions as $\mathcal{D}_{pre}$ and $\mathcal{D}_{ft}$, respectively.
Our model is a linear classifier $W \in \mathbb{R}^{6\times2}$ trained with a cross-entropy loss: %

\begin{equation}
v = Wx, \qquad \hat{P}(y \mid x) = \operatorname{Softmax}(v)_y, \qquad \mathcal{L}_i = -\mathbb{E}_{x,y \sim \mathcal{D}_i}\!\left[\log \hat{P}(y \mid x)\right]
\end{equation}

As shown in Figure~\ref{fig:both}, during the first learning phase (``Pretrained") the model learns a perfect classifier over $D_{pre}$. After the finetuning phase (``Pretrained + FT") it correctly classifies the new distribution $D_{ft}$; however, the decision boundaries from the first phase are heavily deformed, resulting in severe test loss on $\mathcal{D}_{pre}$ - a clear case of catastrophic forgetting.

\noindent \textbf{Global-Support Updates Cause Forgetting.} To better understand this phenomenon we revisit the weight update rule. For training step $s$ and learning rate $\alpha$ we have:
\begin{equation} \label{eq:update_rule}
\Delta W^{(s)}=-\alpha \cdot \frac{d\mathcal{L}}{dW}^{(s)} = - \alpha \cdot \frac{d\mathcal{L}}{dv}^{(s)}x^{(s)T}
\end{equation}
Assume two new test samples $\textcolor{blue}{x_{pre} \in \mathcal{D}_{pre}}$ and $\textcolor{red}{x_{ft} \in \mathcal{D}_{ft}}$, and that the training sample $x^{(s)}$ was seen during the finetuning phase, i.e.\ $\textcolor{red}{x^{(s)} \in \mathcal{D}_{ft}}$. The change in logits for each sample is:

\begin{equation} \label{eq:sample_interaction}
\Delta v_{pre} = \Delta W^{(s)}\textcolor{blue}{x_{pre}} = - \alpha \cdot \frac{d\mathcal{L}}{dv}^{(s)}\textcolor{red}{x^{(s)T}}\textcolor{blue}{x_{pre}}, \:\:\:
\Delta v_{ft} =\Delta W^{(s)}\textcolor{red}{x_{ft}} = - \alpha \cdot \frac{d\mathcal{L}}{dv}^{(s)}\textcolor{red}{x^{(s)T}x_{ft}}
\end{equation}

Notice that the change in logits for {\color{blue}$x_{pre}$} depends on the dot product {\color{red}$x^{(s)T}$}{\color{blue}$x_{pre}$}. While $\Delta W^{(s)}$ is useful for {\color{red}$x_{ft}$}, it is not relevant for most samples in {\color{blue}$\mathcal{D}_{pre}$}. However, unless {\color{blue}$x_{pre}$} is orthogonal to {\color{red}$x^{(s)}$}, this dot product is nonzero, and the update alters the logits of $x_{pre}$ - causing interference. This holds for all updates produced by gradient-based optimizers: Adam rescales gradients by their second moment and Muon orthogonalizes them, but both still produce a matrix update $\Delta W^{(s)}$ that operates on all inputs - making these updates suboptimal.

\noindent \textbf{Local-Support Updates via Support Estimation.}  \citet{saha2021gradientprojectionmemorycontinual} approaches this problem by estimating a linear subspace in which the prior data (here, $\mathcal{D}_{pre}$) lies, and restricting new updates to its orthogonal complement. This works well in small-scale settings, yet becomes infeasible when the prior data includes a full pretraining corpus, which might not even be available.
Therefore, a practical choice is to assume nothing about the prior data - implying that adversarial pretraining samples could exist at any region outside the current data's support.
Concretely, we now diverge from~\citet{saha2021gradientprojectionmemorycontinual} and instead propose to estimate the region where the current data is, denoted $\mathcal{M}$, and restrict the update to that region alone.
As shown in Fig.~\ref{fig:method} (Left), we realize this by placing a gate $ g(x) = x \cdot \mathbb{I} \{x \in \mathcal{M} \}$ before the new phase's weight adapter. Locality is thus controlled by the gate: an update is local if the gate opens only on inputs that are close to the current phase's training data.
Standard finetuning corresponds to an always-open gate, which makes the update global.

\noindent \textit{\textbf{Theoretical Motivation: A density estimator can recover an optimal gate.}} As detailed in App.~\ref{app:theory_full}, against the worst-case test distribution, a gate's error decomposes into two terms, \emph{deficit} and \emph{excess}. The gate's training objective optimizes the deficit directly but is completely blind to the excess. The excess quantifies forgetting: it is the region outside the training distribution on which the gate erroneously stays open. While no objective on the current data alone can act on it, we show that other factors such as the hypothesis class can control it under explicit conditions. We study several hypothesis families, finding that density estimators can recover optimal gates, at costs we analyze.

Following this motivation we solve the toy problem with a local estimator (App.~\ref{app:theory_estimator_fams}): $\mathcal{M}=\bigcup_{i=1}^{N} \mathcal{B}(x_i,r)$, where $\mathcal{B}(x,r)$ is a ball with center $x$ and radius $r$, and $\{x_i\}_{i=1}^{N}$ are the training points from the current phase's data.
In Fig.~\ref{fig:both} (right) we repeat the experiment with this approach, which we call Local Support Learning (LSL - detailed in Sec.~\ref{sec:method}). The LSL-tuned model fully adapts to $\mathcal{D}_{ft}$'s data while having zero interference with $\mathcal{D}_{pre}$, resulting with a perfect classifier. In addition, the decision boundary changed only near the support of $\mathcal{D}_{ft}$, and has not changed elsewhere. \newline
The above gate addresses the problem's core challenge - classifying data from all phases while training only on data from the current one. Yet, it is highly inefficient since its complexity scales with $N$, the number of tokens it encountered. In Sec.~\ref{sec:method}, we propose a much more efficient parametric gate that removes this dependence, with complexity that scales only with the number of phases $P$.

\section{Method}
\label{sec:method}

\noindent \textbf{Gate Design.} We design our gate to be applied individually to each weight matrix. Each gate fits a Gaussian Mixture Model (GMM) to the current phase's input distribution:

\begin{equation} \label{eq:gmm_and_gate_support}
    \Phi(x) = \sum_{k=1}^{K} \pi_k \, \mathcal{N}(x \mid \mu_k, \Sigma_k), \qquad \mathcal{M}=\{x \mid \Phi_{pos}(x) - \Phi_{neg}(x) > 0\}
\end{equation}

where each gaussian $\mathcal{N}$ has mean $\mu_k$, covariance matrix $\Sigma_k$, and mixing weight $\pi_k$. The GMM is fit via the Expectation-Maximization (EM) algorithm.
We compute the gating decision by checking whether the GMM density is above some threshold TH. Conceptually, the support is then $\mathcal{M}=\{x \mid \Phi(x) - \text{TH}>0\}$. This approach requires tuning the threshold, complicating the method. Instead, in Eq.~\ref{eq:gmm_and_gate_support} (right) we propose an input-dependent thresholding approach involving two GMMs.
The first GMM $\Phi_{pos}$ is fit over the current phase's distribution and the second GMM $\Phi_{neg}$ is fit over a \textit{small} sample from a \textit{generic} pre-training dataset.  Now, each input $x$ is classified according to the GMM that gives it a larger likelihood. We emphasize that this reference dataset need not have any relation to the data the model was originally pretrained on, and empirically, does not require consisting more than 1M tokens (for reference, this constitutes about $0.0000067\%$ of Qwen's pretraining set). 

\noindent \textbf{Intuition for $\Phi_{neg}$.}
It may seem unlikely that an exceedingly small, generic,  pretraining sample can capture the support of the full pretraining distribution. The key insight is that the gate need not capture the full support, but only the width of the distribution, which manifests even in a small sample. We illustrate this in 1D in Figure~\ref{fig:method} (Right): Gaussians fit to a narrow distribution $\Phi_{pos}$ have high likelihood only in their vicinity, while the wider $\Phi_{neg}$ dominates elsewhere.
In high dimensions, $\Phi_{\mathrm{neg}}$ need not be wider than $\Phi_{\mathrm{pos}}$ in every direction, and constructing an ideal threshold, one that is exceeded exactly on the current phase's data, is not trivial.
$\Phi_{\mathrm{neg}}$ is a practical approximation: the gate still opens on about 16\% of tokens from out-of-distribution pretraining tasks (5\% with temporal smoothing), yet retains 96.6\% of pretrained performance (98.8\% with smoothing; Fig.~\ref{fig:gmm_ablation_ncomps_temp_smooth}).

\vspace{-14pt}

\begin{figure}[H]
\begin{minipage}[t]{0.48\textwidth}
\begin{algorithm}[H]
\caption{LSL: Training} \label{alg:msl_training}
\begin{algorithmic}[1]
\Require Model $M$, datasets $\{\mathcal{D}_{i}\}_{i=1}^{P}$, num epochs $E$, reference data $\mathcal{D}_{\text{ref}}$
\Ensure $M \oplus \bigcup_{i=1}^{P} \{\mathbf{W}_{\text{adapter}}^{(i)}, \boldsymbol{\Phi}_{pos}^{(i)}, \boldsymbol{\Phi}_{neg}\}$
\Statex $\triangleright$ $\oplus$ denotes mounting onto the base model
\State $\boldsymbol{\Phi}_{neg} \gets \Call{FitGMMs}{M, \mathcal{D}_{\text{ref}}}$
\For{$i = 1, \dots, P$}
    \State $\mathbf{W}_{\text{adapter}}^{(i)} \gets \Call{FitAdapters}{M, \mathcal{D}_{i}, E}$
    \State $\boldsymbol{\Phi}_{pos}^{(i)} \gets \Call{FitGMMs}{M \oplus \mathbf{W}_{\text{adapter}}^{(i)}, \mathcal{D}_{i}}$
    \State $M \gets M \oplus \{\mathbf{W}_{\text{adapter}}^{(i)}, \boldsymbol{\Phi}_{pos}^{(i)}, \boldsymbol{\Phi}_{neg}\}$
\EndFor
\State \Return $M$ %
\end{algorithmic}
\end{algorithm}

\end{minipage}
\hfill
\begin{minipage}[t]{0.48\textwidth}
\begin{algorithm}[H]
\caption{LSL: Inference (Per-Module)} \label{alg:msl_inference}
\begin{algorithmic}[1]
\Require Input $x$, model $M$, smoothing states $\{s^{(i)}\}_{i=1}^{p}$ and coefficient $\alpha$.
\Statex $\triangleright$ each $s^{(i)}$ initialized to $0$ if not set.
\State $v \gets W_{\text{pretrained}} \, x$
\For{$i = 1, \dots, p$}
    \State $g \gets \mathbb{I}\{\boldsymbol{\Phi}_{pos}^{(i)}(x) - \boldsymbol{\Phi}_{neg}(x) > 0\}$
    \State $s^{(i)} \gets \alpha \cdot g + (1 - \alpha) \cdot s^{(i)}$
    \State $v \gets v + \mathbb{I}\{s^{(i)} > 0.5\} \cdot W_{\text{adapter}}^{(i)} \, x$
\EndFor
\State \Return $v$, $\{s^{(i)}\}_{i=1}^{p}$
\end{algorithmic}
\end{algorithm}
\end{minipage}
\vspace{-4pt}
\caption{\textbf{Local Support Learning.} We describe the training and inference algorithms. Inference is presented per-module (a weight matrix in some layer) and training is presented for the full model. See Alg.~\ref{alg:fit_gmm} (FitGMMs) and Alg.~\ref{alg:fit_adapter} (FitAdapters) for full details.  
$\mathbb{I}\{\cdot\}$ is the indicator function.}
\end{figure}

\noindent \textbf{Temporal Smoothing.} The GMM gates each token independently, with no notion of temporal continuity. We address this with a simple exponential moving average that smooths decisions over time, leveraging the fact that task boundaries do not change at a per-token frequency. Importantly, the core mechanism accounting for the vast majority of retention is the GMM, while the smoothing aggregator is only a lightweight complement. We support this with detailed ablation studies showing that the GMM improves retention from 76.6\% to 96.6\%, and the aggregator improves retention further from 96.6\% to 98.8\% (Fig.~\ref{fig:gmm_ablation_ncomps_temp_smooth} and Sec.~\ref{sec:ablations}).

\noindent \textbf{Local Support Learning.}
The full training algorithm can be found in Alg.~\ref{alg:msl_training}. We first fit $\Phi_{neg}$. Then, in each phase $i$ we fit the $i$-th weight adapter via gradient descent, followed by fitting its respective $\Phi_{pos}$ GMM. Notice that adapters and GMMs from previous phases are active during the current training phase. While FitGMMs (Alg.~\ref{alg:fit_gmm}) is straightforward algorithmically, it introduces some technical challenges, addressed in Appendix~\ref{sec:gmm_fitting_algo}. FitAdapters (Alg.~\ref{alg:fit_adapter}) is identical to standard adapter finetuning (e.g., LoRA); additional design choices are discussed in the same appendix. \newline
The inference algorithm is provided in Alg.~\ref{alg:msl_inference}, presented per-token. Importantly, it is fully parallelizable during prefill, just like a standard MLPs. We first compute the output for the pretrained weights. Then, for each previously seen phase $i$, we compute gate $i$'s decision and apply temporal smoothing. The routing decision of adapter $i$ is based on the smoothed value. See complexity analysis in App.~\ref{sec:compl_acc_extended}

\vspace{-4pt}

\section{Experimental Results}

\subsection{Post-Training with Local Support Learning}
\label{sec:experiments_msl_llms}
\vspace{-6pt}
In this section we thoroughly evaluate our approach. Implementation details can be found in App.~\ref{sec:impl_details}.

\begin{figure}[h]
  \centering
  \includegraphics[width=0.32\textwidth]{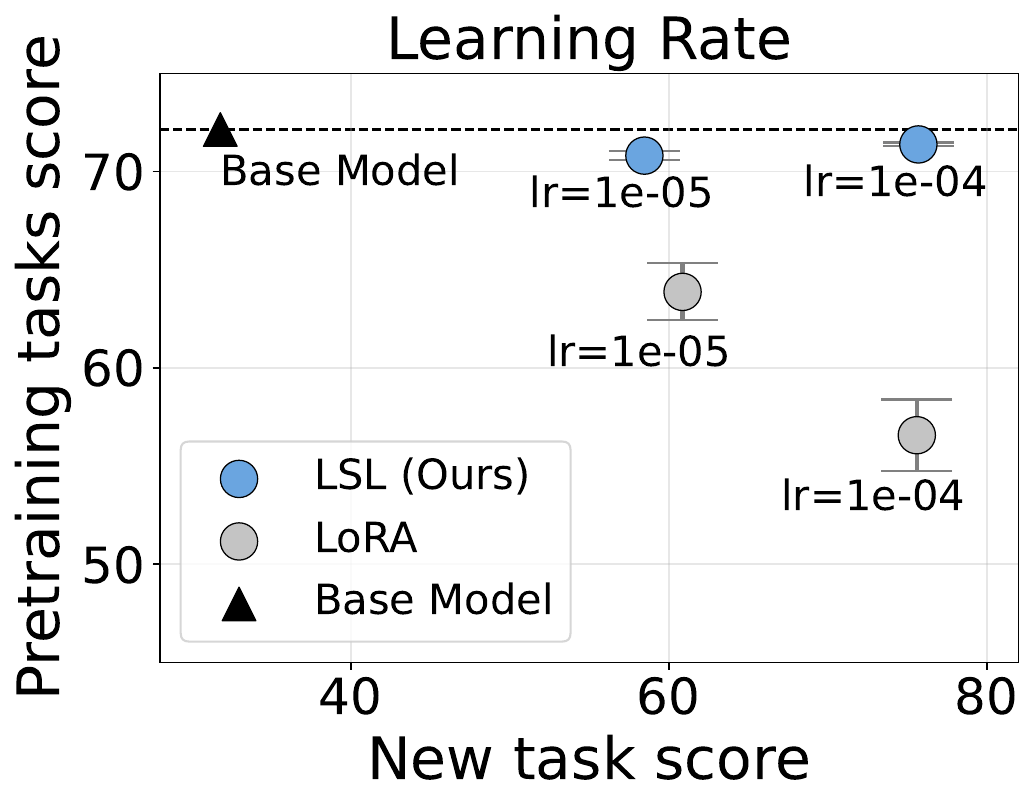}
  \includegraphics[width=0.32\textwidth]{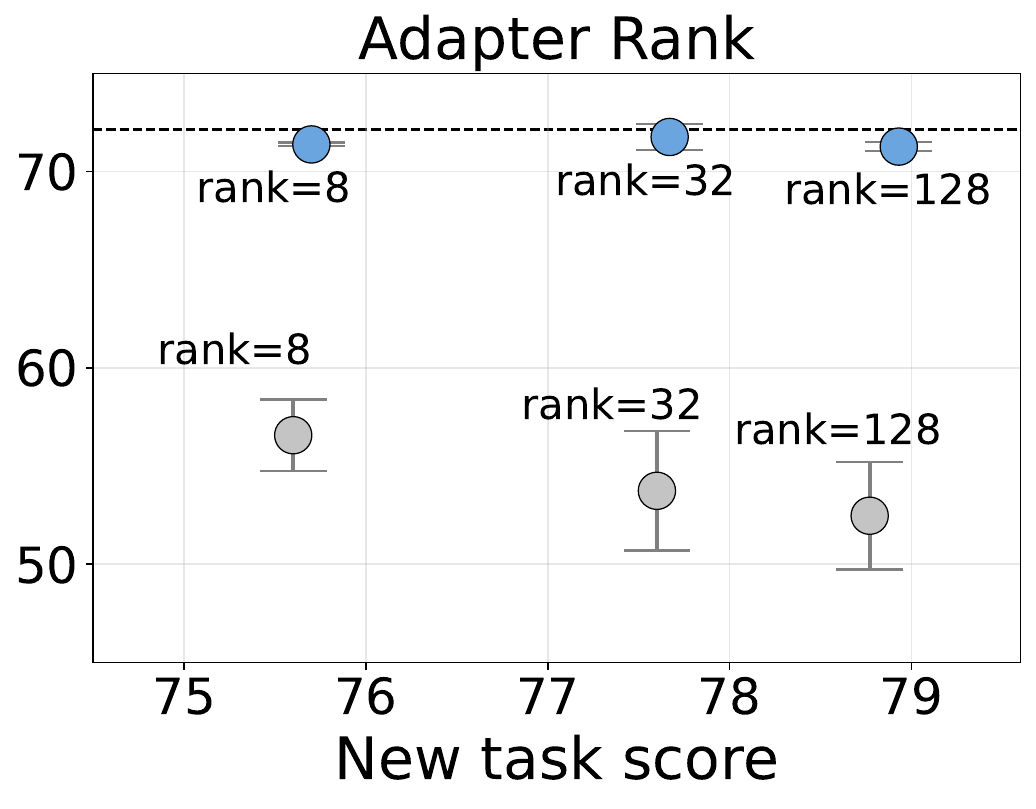}
  \includegraphics[width=0.32\textwidth]{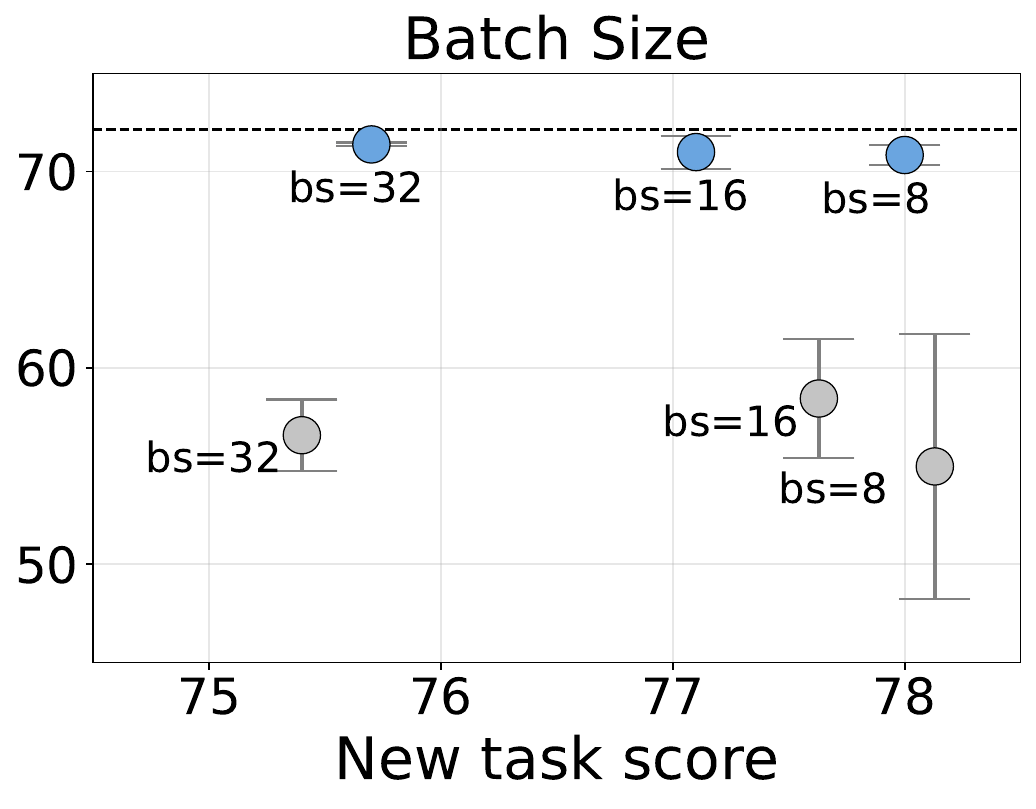}
  \vspace{-8pt}
  \caption{\textbf{Robustness to hyperparameters}. In each plot, we sweep one of the hyperparameters. While conventional fine-tuning trades off learning against forgetting, our method decouples the two, allowing the model to reach its peak capability.}
  \label{fig:cl_llm_rank}
  \vspace{-6pt}
\end{figure}

\noindent \textbf{Baselines.} We finetune Qwen2.5-7B-Instruct~\citep{qwen2025qwen25technicalreport}, a strong base model with established math, coding, and instruction-following capabilities.
We compare LSL against two weight-based retention techniques: (i)~LoRA~\citep{hu2021loralowrankadaptationlarge} which constrains updates to a low-dimensional subspace, and (ii)~OP-LoRA~\citep{xiong2025oploraorthogonalprojectionlora}, 
which minimizes interference between the update and the dominant singular directions of the pretrained weights. We also examine one regularization-based approach: (iii) Learning without Forgetting (LwF)~\citep{li2017learningforgetting}, which penalizes change in the model's output relative to the base model, using only finetuning data. We exclude methods that do not target retention of pretraining capabilities, such as O-LoRA, and methods that are impractical in pretraining settings, such as replay buffers (see Sec.~\ref{sec:rw} and App.~\ref{appendix:extended_rw}). In the hyperparameter sweep, we optimize for new-task validation performance; retention benchmarks are never used for selection. This compares all methods at their full learning capacity, where the trade-off between learning and forgetting is most pronounced (Fig.~\ref{fig:cl_llm_rank})

\noindent \textbf{Finetuning tasks.} We evaluate three post-training settings: (i)~cybersecurity instruction tuning~\citep{trendyol_2025_cybersec_v2}, (ii)~low-resource translation from English to Igbo~\citep{nwafor-nguyen-2025-fostering}, and (iii)~chemistry instruction tuning~\citep{zhang2024chemllmchemicallargelanguage}.
These tasks are chosen to be diverse and challenging, testing not only retention of pretrained capabilities but also sufficient adaptation to each new domain: English$\rightarrow$Igbo is an open-ended generation task into a low-resource language; cybersecurity consists of instruction-response pairs requiring specialized domain knowledge; and chemistry demands understanding of molecular structure and interactions, expressed via a blend of natural language and SMILES molecular notation.
Samples from each dataset are provided in Fig.~\ref{fig:dataset_samples}. English$\rightarrow$Igbo translation is evaluated using chrF~\citep{popovic-2015-chrf}, a character-level metric suited for low-resource languages. Chemistry and cybersecurity are evaluated on ChemBench~\citep{zhang2024chemllmchemicallargelanguage} and SecEval~\citep{li2023seceval}, respectively, both multi-choice QA tasks.

\begin{figure}[h]
  \centering
  \includegraphics[width=\textwidth]{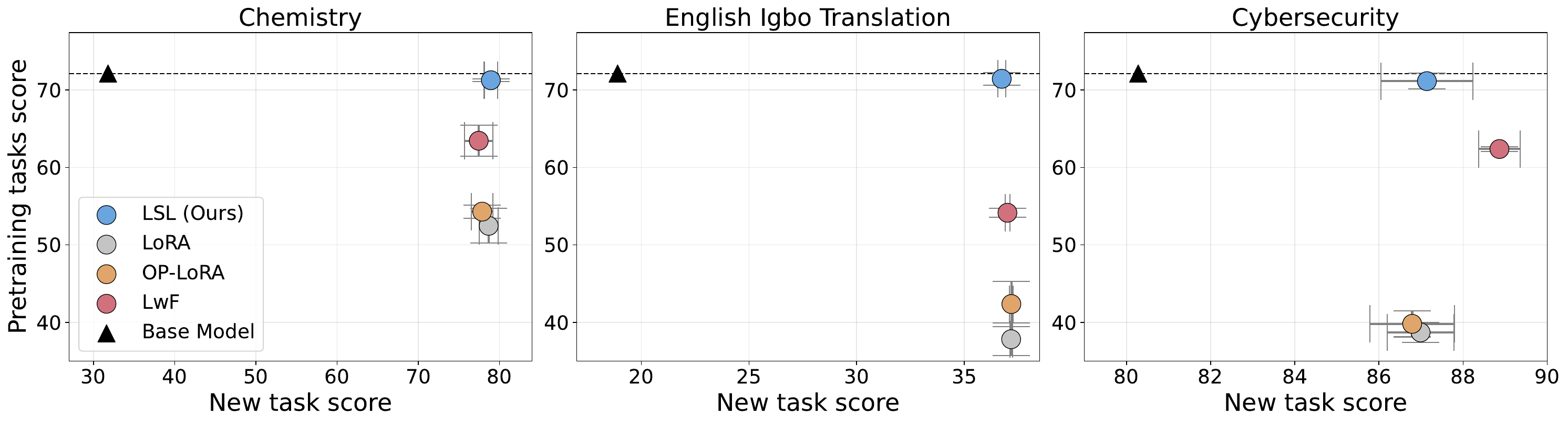}
  \vspace{-16pt}
  \caption{\textbf{Post-training with LSL}. We evaluate forgetting on a Qwen2.5-7B-Instruct model for three diverse downstream tasks. The $x$-axis measures the new task's performance and the $y$-axis measures capability retention - the average score across three pretraining benchmarks. The upper-right corner reflects stronger performance. LSL learns the downstream task while achieving near-optimal retention across all settings. This is in contrast to the baselines, which apply new updates to all inputs and thus are prone to interference on prior task data.}
  \label{fig:cl_llm}
  \vspace{-14pt}
\end{figure}

\noindent \textbf{Pretrained capability retention.} To measure forgetting, we evaluate on three diverse tasks reflecting the base model's capabilities: GSM8K for math reasoning~\citep{cobbe2021trainingverifierssolvemath}, HumanEval for code generation~\citep{chen2021evaluatinglargelanguagemodels}, and IFEval for instruction following~\citep{zhou2023instructionfollowingevaluationlargelanguage}. \newline
Our results are reported in Fig.~\ref{fig:cl_llm}. Across all benchmarks, LSL achieves strong performance on the new task while maintaining close to optimal retention.
LoRA exhibits strong adaptation, yet suffers from forgetting in all settings.
OP-LoRA performs very similarly to the LoRA baseline. We find that this is because forgetting is only weakly tied to the top-$k$ singular-value subspace of the pretrained weights (see App.~\ref{app:op_lora_performance}).
The function-regularization technique, LwF, performs better than the weight-based retention techniques, yet still falls behind LSL across all tasks. Common to all baselines is that their gradient updates are applied to all inputs, which, according to the analysis in Sec.~\ref{sec:prob_inv}, leaves them prone to interference with prior data. We study this in detail in the ablations section (Sec.~\ref{sec:ablations}).

\begin{figure}[h]
  \centering
  \includegraphics[width=0.9\textwidth]{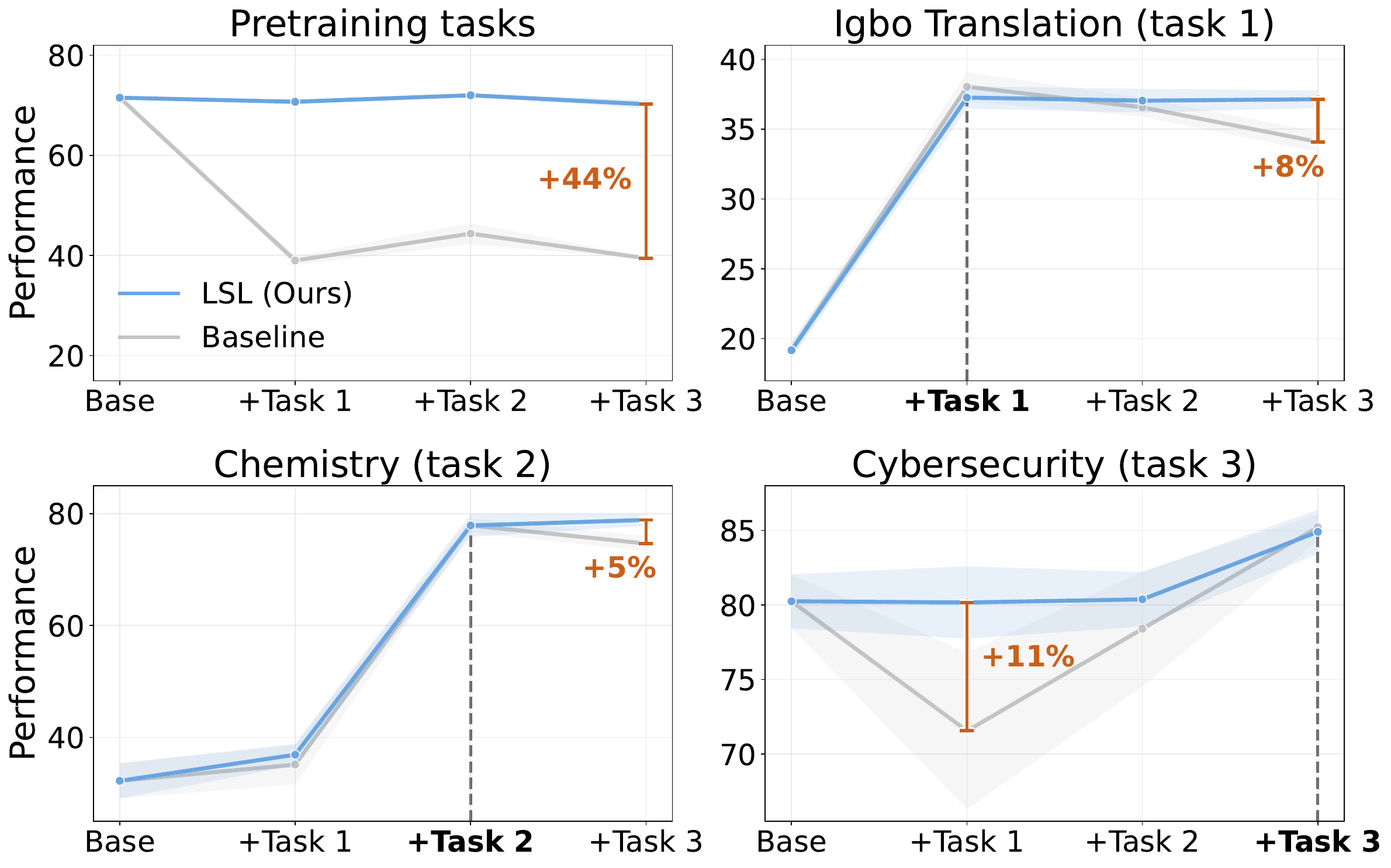}
  \vspace{-3pt}
  \caption{\textbf{Post-Training with multiple phases}. 
  Each plot tracks the evaluation score of one task throughout training; the phase whose training data matches the evaluated task is highlighted in bold. The training order is (1) Igbo translation, (2) Chemistry, (3) Cybersecurity. In the leftmost plot, LSL retains pretrained capabilities throughout, while the baseline drops by 44\% already after one phase. On the three finetuning benchmarks, LSL retains its gains after each task's phase ends, whereas the baseline's performance declines steadily once the peak is reached.}
  \label{fig:cl_llm_multiple_phases}
  \vspace{-9pt}
\end{figure}

\noindent \textbf{Multiple Post-Training Phases.} In Fig.~\ref{fig:cl_llm_multiple_phases}, the model is trained on all finetuning tasks sequentially. We compare LSL to a LoRA baseline that allocates a new adapter per phase and merges it into the model before the next phase begins. In the leftmost plot, LSL retains pretrained capabilities across all phases, whereas the baseline degrades severely already after the first phase. The other plots show performance on each finetuning task. With LSL, performance on a task is preserved after its training phase ends, showing that interference between different finetuning phases is also avoided. In contrast, LoRA forgets once the corresponding phase ends: task 1 (translation) drops by 8\% after two phases, and task 2 (chemistry) by 5\% after one. Notably, after task 1, the zero-shot performance on task 3 (cybersecurity) falls by 11\% - a form of forgetting that LSL again avoids. 

\vspace{-8pt}

\noindent\textbf{Robustness to Hyperparameters.}
By avoiding interference, LSL naturally decouples learning from forgetting. A direct benefit is that learning hyperparameters - traditionally considered to trade off with retention~\citep{biderman2024loralearnsforgets} - no longer affect forgetting. In Fig.~\ref{fig:cl_llm_rank} we sweep the learning rate, adapter rank, and batch size for both LSL and LoRA. LSL learns at full capacity without sacrificing pretraining capabilities across all sweeps, while the traditional approach faces the aforementioned trade-off, limiting the model from reaching its full potential.

\begin{wrapfigure}{r}{0.38\textwidth}
  \centering
  \includegraphics[width=0.37\textwidth]{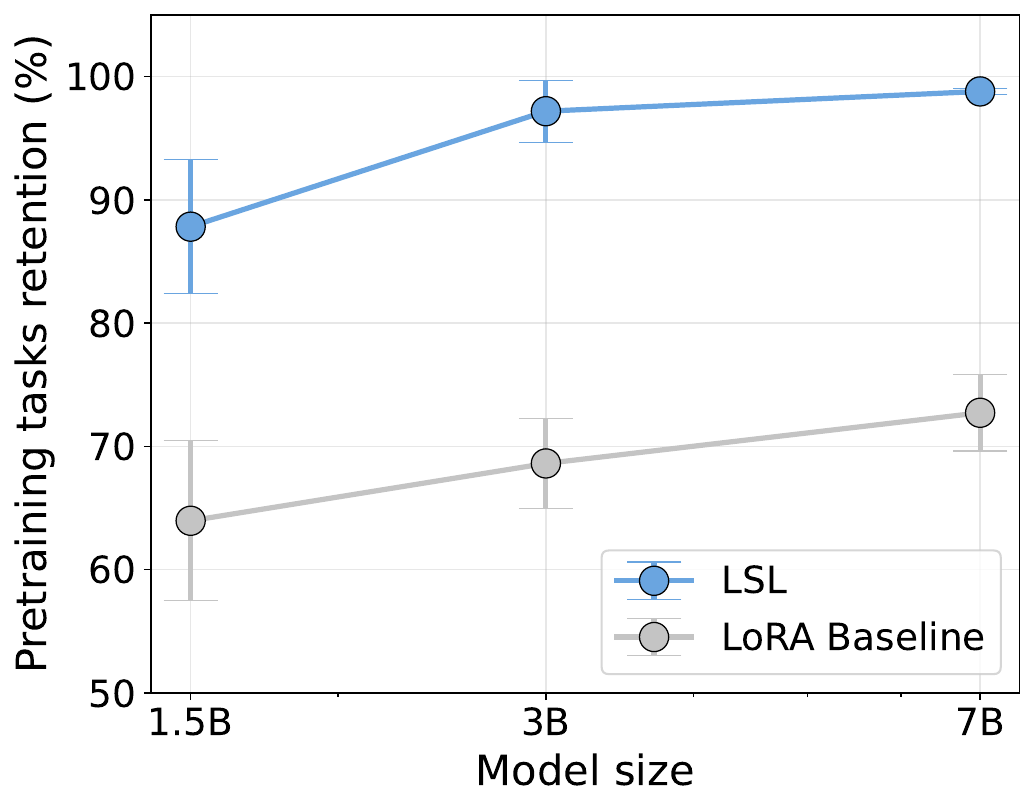}
  \vspace{-12pt}
  \caption{\textbf{Scaling behavior.} We measure the retention of pretraining tasks, defined as the ratio of performance before and after finetuning. LSL exhibits desirable scaling trends.}
  \label{fig:scaling_experiment}
  \vspace{-20pt}
\end{wrapfigure}

\noindent\textbf{Scaling Behavior.} In Fig.~\ref{fig:scaling_experiment} we test LSL across different model sizes. Each point shows the retention of pretraining tasks, defined as the ratio of performance before and after finetuning. LSL works at all scales tested (1.5B to 7B), and its performance improves with scale. LoRA also improves with scale, but far more slowly, remaining well below LSL at every size. This highlights LSL's scaling potential.

\begin{figure}[h]
  \centering
  \includegraphics[width=\textwidth]{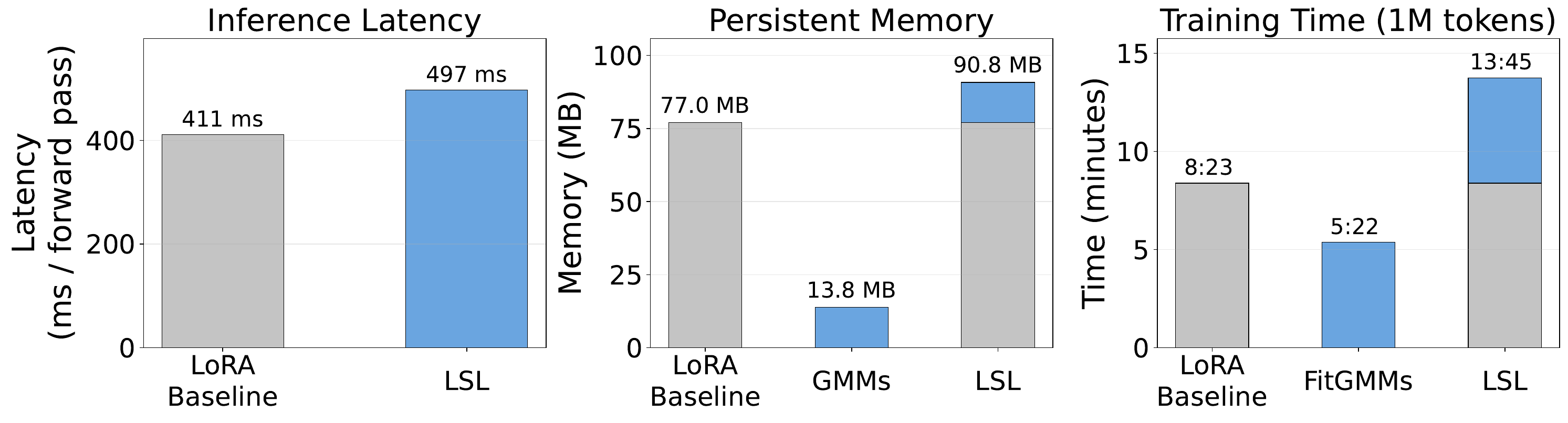}
  \vspace{-20pt}
  \caption{\textbf{Efficiency benchmarks} on Nvidia B200 with Qwen2.5-7B-Instruct. LSL achieves strong retention with memory, inference, and training overhead on par with established baselines that do not achieve comparable retention.}
  \label{fig:cl_llm_efficiency}
  \vspace{-4pt}
\end{figure}

\noindent \textbf{Efficiency}
In Fig.~\ref{fig:cl_llm_efficiency} we benchmark LSL on a Nvidia B200 GPU with the Qwen2.5-7B-Instruct model. We highlight three practical results. First, LSL requires only 13.8MB of additional memory for the GMMs ($\Phi_{pos}$, $\Phi_{neg}$). Since $\Phi_{neg}$ is shared, this results with a negligible overhead of 6.9MB per learning phase. Second, a LSL adapter adds only 86ms of inference latency per forward pass, already competitive with standard LoRA baselines. Third, adapter training time on 1M tokens is comparable to a LoRA baseline; adding the GMM fitting step increases the total training time by $\times1.64$. We exclude $\Phi_{neg}$ fitting as it is a one-time cost independent of the finetuning data. Note that neither inference nor training were fully optimized. The key takeaway is that LSL's overhead is on par with established baselines, while achieving superior retention.

\subsection{Ablations}
\label{sec:ablations}

\begin{figure}[h]
  \centering
  \includegraphics[width=0.608\textwidth]{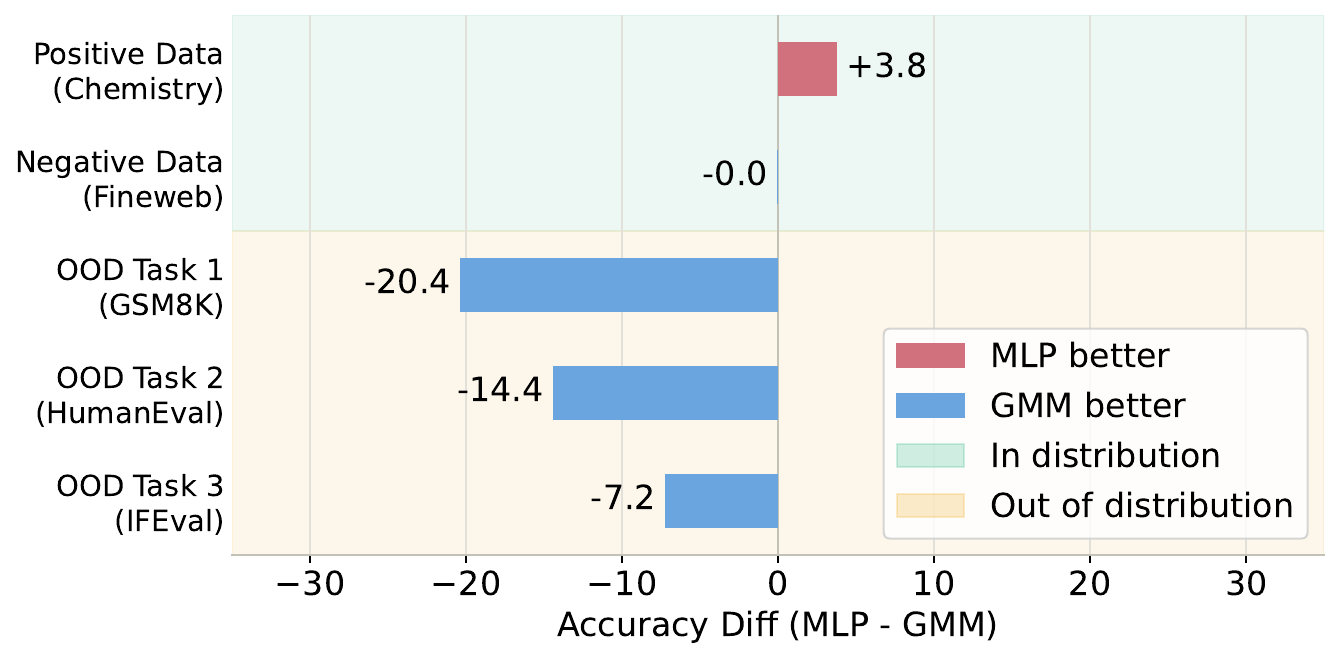}
  \includegraphics[width=0.38\textwidth]{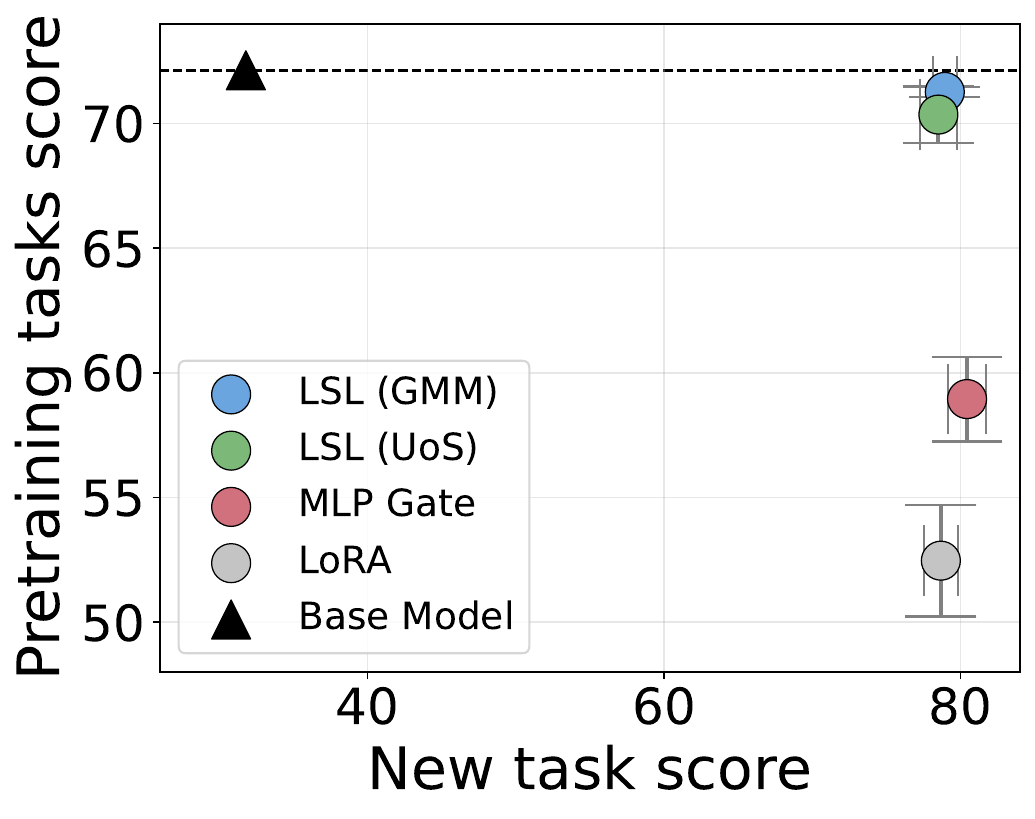}
  \vspace{-4pt}
  \caption{\textbf{Gate ablations.} (\textbf{Left}) Classification accuracy of the MLP and GMM gates at an intermediate layer. The GMM tends to stay closed on data from previous learning phases, a property that enables stronger retention. (\textbf{Right}) Retention gains stem from support locality rather than a specific architecture. Replacing the GMM with a different local gate, Union of Spheres (UoS, green), yields equivalent performance, whereas a non-local MLP gate (red) degrades retention significantly.}

  \label{fig:gmm_vs_uos}
  \vspace{-10pt}
\end{figure}

\noindent \textbf{Importance of Locality.} An important question is whether the retention gains stem from the GMM itself, or from a more general property - local support of the update. To disentangle the two, we construct two alternative support estimators and use them to ablate the GMM. The first is Union of Spheres (UoS), defined as $\mathcal{M}=\bigcup_{i=1}^{M} \mathcal{B}(x_i,r)$, where $\mathcal{B}(x,r)$ is a ball with center $x$ and radius $r$, and $\{x_i\}_{i=1}^{M}$ are representative points from the current phase's data. By construction, this gate's support is local (App. ~\ref{app:theory_estimator_fams}). The gate is optimized greedily: each new point encountered during training is discarded if it falls within $\mathcal{M}$, and added to the representative set otherwise. The radius $r$ is selected per gate via a calibration process detailed in Appendix~\ref{sec:uos_r_calib}. The second is an MLP classifier trained on the same data used to fit the GMM ($\Phi_{\mathrm{pos}}$ and $\Phi_{\mathrm{neg}}$). Importantly, the MLP's hypothesis class does not guarantee that the gate will be closed outside its training data (App. ~\ref{app:theory_estimator_fams}).
In Fig.~\ref{fig:gmm_vs_uos} (right), we repeat the experiment from Sec.~\ref{sec:experiments_msl_llms} with the new gates. The GMM and UoS gates achieve close to optimal retention scores while learning at full capacity. In contrast, the MLP lags behind in retention. This supports our hypothesis that local-support updates are the key ingredient for retention, rather than a specific architecture.
The main difference between the UoS and GMM gates is efficiency: the UoS gate is a lookup table that stores many samples, requiring about 5.67GB of memory. In contrast, the parametric GMM learns the support with far less memory - 13.8MB in total, comparable to the LoRA adapter. For reference, the base model weights require about 14GB (in bfloat16), making the GMM overhead negligible - three orders of magnitude smaller.

\noindent \textbf{Gate Behavior on OOD Data.} Since the gate is unaware of prior learning phases, it must stay closed when encountering out-of-distribution (OOD) data. In Fig.~\ref{fig:gmm_vs_uos} (Left) we compare the classification accuracy of the GMM and MLP gates. The MLP has a slight edge on in-distribution data, yet it degrades significantly on all three OOD pretraining tasks. This further confirms that the update's locality to its training data, rather than in-domain expressiveness, is key for retention(App.~\ref{app:theory_full}), suggesting that the gate's behavior on OOD data is the main driver for the performance in Sec.~\ref{sec:experiments_msl_llms}.

\begin{figure}[h]
  \centering
  \includegraphics[width=\textwidth]{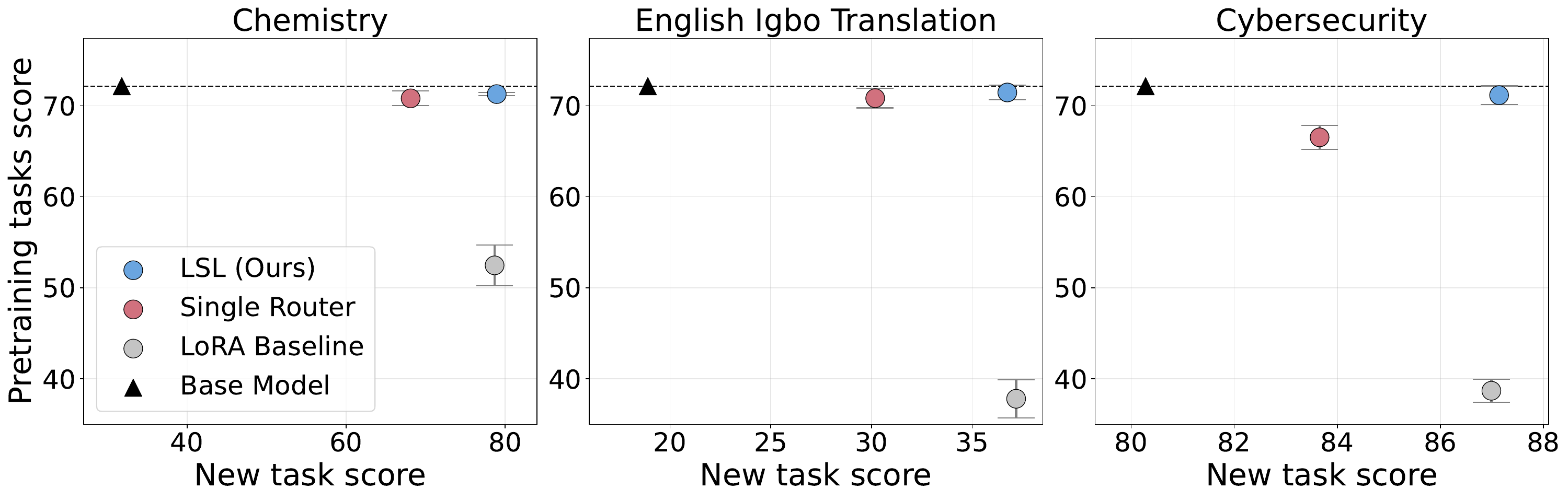}
  \vspace{-20pt}
  \caption{\textbf{Single router ablation.} We replace LSL's per-matrix gates with a single gate after the embedding layer, whose decision controls all adapters. The single router retains far more than LoRA but falls short of LSL, especially on the new task. It also degrades retention on cybersecurity. Per-matrix gates decide from contextual features at every depth, allowing full learning and retention.}
  \label{fig:single_router_ablation}
  \vspace{-4pt}
\end{figure}

\noindent \textbf{Why per-matrix gates?} One might expect a single routing decision per token to suffice, since a token either belongs to the current phase or it does not. To test this, we replace LSL's per-matrix gates with a single per-token gate placed after the embedding layer, whose decision controls all adapters (Fig.~\ref{fig:single_router_ablation}). The single router recovers only 50-77\% of LSL's improvement on the new task and, on cybersecurity, also degrades retention. The main trend can be attributed to surface-level features identifying fewer new-task tokens, leading to degraded new-task performance. In addition, the single-router's decision propagates an error to every adapter. Per-matrix gates instead decide from contextual features at every depth, and an error at one gate affects only its own matrix. For a broader discussion including other alternatives see App.~\ref{app:single_router}.

\noindent \textbf{Number of GMM Components and Temporal Smoothing}
In Fig.~\ref{fig:gmm_ablation_ncomps_temp_smooth} (Appendix) we ablate the remaining components of our method. We find that LSL is largely insensitive to the number of GMM components - achieving near-optimal retention with as few as two - highlighting the quality of the base model's embeddings. Temporal smoothing further improves retention from a score of 69.8 to 71.1 (96.6\% to 98.8\% with respect to the base model's performance), primarily by further reducing the hit rate on pretraining tasks. Yet, the dominant contribution comes from the GMM, which improves retention from 55.3 to 69.8 (76.6\% to 96.6\% with respect to the base model).

\noindent \textbf{Function Behavior Change} To measure the gate's impact directly in function space, we compare the per-token output distributions of the finetuned and base models, grouped by the fraction of LSL gates each token opens (App.~\ref{sec:fbc}). We find that on in-distribution data, LSL and LoRA change the model's behavior similarly, while on out-of-distribution pretraining benchmarks, LSL changes behavior much less.

\newcommand{\strike}[1]{%
  \setbox0=\hbox{#1}%
  \rlap{\hbox to \wd0{\hfil\vrule height 0.5ex depth -0.4ex width \wd0\hfil}}%
  \box0%
}

\section{Conclusion}

We studied catastrophic forgetting in large pretrained models through the geometry of each weight matrix's input space. This view yields a natural retention objective: updates should be restricted to activations from their training distribution. This reveals that updates produced by gradient-based optimizers are suboptimal, since the matrix update they produce acts on all possible inputs. Building on this analysis, we proposed Local Support Learning, which pairs a standard adapter with a gating function fit by a tailored Gaussian Mixture Model. The gate's local support is what allows it to route inputs from all phases while training only on the current one. On LLMs of up to 7 billion parameters, LSL learns at full capacity while retaining pretrained capabilities and previously finetuned skills across multiple sequential phases. It is also robust to hyperparameter choice, incurs a small memory footprint independent of the number of training tokens, and shows promising scaling behavior.

\textbf{Limitations.} LSL's adapters cannot be merged into the base weights, adding inference cost that grows linearly with the number of phases $P$. However, this computation is highly parallelizable, allowing to trade compute for inference time. Note that $P$ counts phases, not tasks - many tasks can be learned jointly within a phase. In addition, our theoretical motivation concerns a single weight matrix; its extension to the full multi-layer network is supported empirically but not formally. Lastly, LSL relies on the quality of the base model's representations, since the gate assumes each phase's activations can be captured by a GMM. This is reflected in the scaling experiment, where larger models achieve better retention (Fig.~\ref{fig:scaling_experiment}).

\textbf{Future research} includes extending LSL to larger models, other modalities, reinforcement learning objectives, and hundreds of phases. The latter would open the door to test-time training, a form of learning currently avoided in practice due to the unpredictability of forgetting, with the potential to store new knowledge in the weights rather than in context, at a fraction of the cost. Lastly, LSL itself can be further explored, e.g., through other local gating functions and improved acceleration.

\subsection*{AI use statement}
In this work, we used generative AI tools for code implementation, for identifying additional related work, for checking the fluency and formatting of the manuscript, and for refining mathematical claims and proofs. We have not used generative AI tools for generating data or running experiments. We have reviewed all AI-assisted work. Code was written with an AI coding assistant in a mode where every proposed edit was reviewed and
approved by an author before being applied, and its correctness was verified by the authors as well as through internal unit testing. Related work surfaced by
AI tools was read and verified by the authors before citation. Manuscript suggestions were limited to wording and formatting.  We take responsibility for the final content of this work, including text, claims, or
artifacts produced with the aid of generative AI.

\subsection*{Ethics statement}

This work analyzes and proposes a method to overcome catastrophic forgetting, which is crucial when deploying LLMs in real-world systems. This improvement is anticipated to have a positive impact on the use of LLMs in society, particularly by increasing model efficiency and reducing energy consumption. However, we acknowledge that continual learning methods, including ours, inherit the limitations of the underlying pretrained model and training data. We emphasize the necessity of careful evaluation before deployment beyond the research environment.

\subsection*{Reproducibility statement}

First, we provide the source code used for the key experiments. Second, in Appendix~\ref{sec:impl_details} we provide full configurations for all experiments including instructions on how to train and evaluate the models. We also explicitly state all used datasets, models and hardware, and describe the exact metrics used in each experiment.

\subsubsection*{Acknowledgments}
{This work was supported by a grant from the Tel Aviv University Center for AI and Data
Science (TAD). This research was made possible through a GPU compute resource grant funded by the Association of University Heads, the Council for Higher Education and the AI Research Compute Center. We thank Ryan Bahlous-Boldi and Idan Shenfeld for insightful discussions and helpful feedback that improved this work.}

\bibliography{iclr2027_conference}
\bibliographystyle{iclr2027_conference}

\appendix

\newpage

\section{Extended Related Work}
\label{appendix:extended_rw}

Replay-based methods mix pretraining-like data into the finetuning set, whether from external corpora \citep{scialom2022finetunedlanguagemodelscontinual}, generated by the model itself \citep{huang2024mitigatingcatastrophicforgettinglarge}, or routed to dedicated adapters \citep{dou2024loramoealleviateworldknowledge} that re-learn information that may be forgotten; however, approximating the pretraining distribution during fine-tuning is impractical, and these methods can only protect capabilities represented in the replayed samples. \newline
A recent line of work has shown that on-policy reinforcement learning is able to mitigate forgetting of pretraining capabilities \citep{shenfeld2025rlsrazoronlinereinforcement, shenfeld2026selfdistillationenablescontinuallearning, chen2026retainingdoingroleonpolicy}. These methods show promise, yet are confined to RL-based settings, rely on in-context learning capabilities and necessitate a full rollout at each learning step - requirements that our method does not impose. Furthermore, in those settings, these methods are complementary to ours since they target the training objective and data rather than the architecture and optimizer. \newline
In lifelong model editing, several works restrict new parameters to the activations they were trained on.  GRACE \citep{hartvigsen2023aginggracelifelongmodel} caches layer activations as keys with a deferral radius and applies an edit only when an input falls within it, a local-support gate closely related to our Union-of-Spheres ablation (Sec.~\ref{sec:ablations}), while WISE \citep{wang2024wiserethinkingknowledgememory} routes tokens between a side memory and the original weights by thresholding an activation score. Both target the editing of individual facts at a single chosen layer. LSL, in contrast, is a general-purpose learning framework: it trains the whole model on full finetuning datasets.

\begin{figure}[h]
  \centering
  \includegraphics[width=\textwidth]{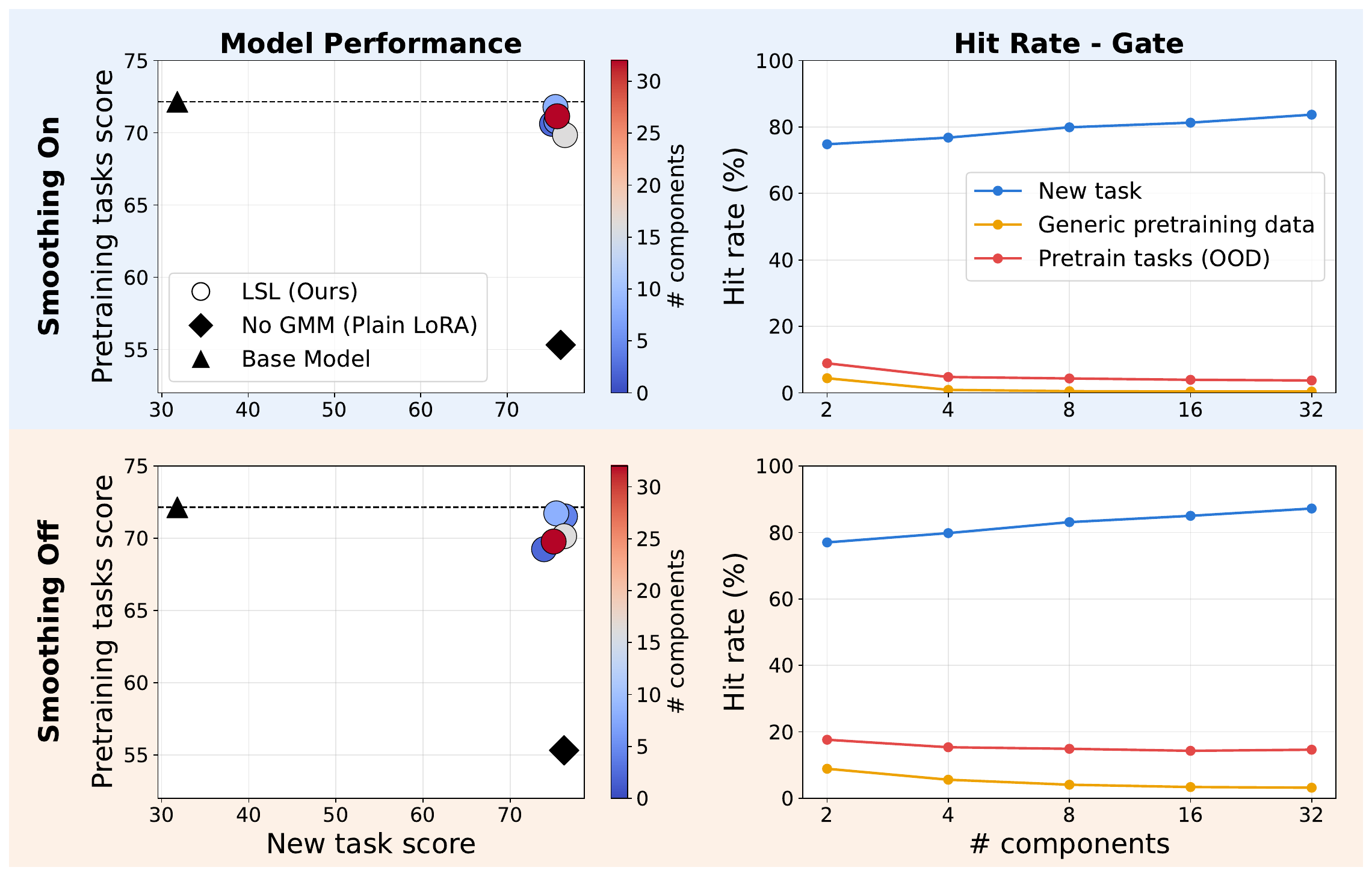}
  \caption{\textbf{Ablating the Number of GMM Components and Temporal Smoothing}. {We measure performance via overall model performance (left column) and average hit rate across gates (right column). An optimal gate achieves 100\% hit rate on the new task (blue) and 0\% on pretrain data (yellow, red). Note that each plot sweeps the number of GMM components. We find that LSL is largely insensitive to the number of GMM components. In addition, the GMM accounts for the vast majority of retention, while temporal smoothing adds the final increment, mainly by further reducing the hit rate on pretraining tasks.}}
  \label{fig:gmm_ablation_ncomps_temp_smooth}
\end{figure}

\section{Complexity and Acceleration} \label{sec:compl_acc_extended}
We analyze the overhead relative to typical adapter-based finetuning. LSL adds one GMM per weight matrix per phase; a naive implementation requires $O(PLKD^2)$ persistent memory and time complexity of $O(PBLKD^2)$ per forward-pass. This is for $P$ phases, $K$ GMM components, $L$ layers, hidden dimension $D$, and batch size $B$. Crucially, the memory cost per phase is independent of the number of tokens, unlike replay buffers or in-context memory, which grow with the data. Inference time, however, is also linear in $P$: unlike traditional finetuning, the adapters cannot be merged into the weights. 
The naive implementation's $D^2$ dependence is undesirable and hinders scaling. We alleviate it through two key accelerations that reduce the memory complexity to $O(PLKE+LDE)$ and time complexity to $O(PBL(DE+KE))$, where $E \ll D$. These are achieved by: (1) applying the Johnson--Lindenstrauss lemma to compress gate inputs to dimension $E$, and (2) using diagonal covariance GMMs, which we find sufficient. 
Training cost of the GMMs is comparable to typical adapter-based finetuning. In addition, the GMM fitting step is lightweight (Expectation-Maximization), but scaling to large finetuning datasets may require further acceleration. Another practical improvement is sharing gates across matrices with identical inputs (e.g., $W_q, W_k, W_v$). This does not change the asymptotic complexity, yet reduces the number of gates by 43\% in a standard Qwen/LLaMA architecture. Also, since different gates train independently, parallelizing their training is highly straightforward.
GMM training, as detailed in Alg.~\ref{alg:fit_gmm}, is performed module-by-module; therefore, beyond the typical complexity of fitting each GMM, the only overhead is a single forward pass to generate the inputs---which is completely negligible.
Empirical efficiency evaluations are in Sec.~\ref{sec:experiments_msl_llms}.

\section{Explaining OP-LoRA's Performance}
\label{app:op_lora_performance}

In Sec.~\ref{sec:experiments_msl_llms} we found that OP-LoRA performs very similarly to the LoRA baseline.
To understand why, we refer to  Table~\ref{tab:op_lora_rho}, which reports the $\rho_k$ metric from the OP-LoRA paper - measuring how much energy of a standard LoRA adapter falls within the subspace of the top-$k$ singular values of the pretrained weights. 

\begin{table}[h] 
\centering
\begin{tabular}{c|cccccc}
 $k$ & 512 & 128 & 64 & 16 & 4 & 1 \\
\hline
$\rho_k$ & 1.70\% & 0.44\% & 0.27\% & 0.12\% & 0.06\% & 0.02\%
\end{tabular}
\caption{\textbf{Evaluating the effectiveness of OP-LoRA.} We report the $\rho_k$ metric from the OP-LoRA paper, which measures how much
energy of a standard LoRA adapter falls within the subspace of the top-$k$ singular values of the pretrained weights (trained on the chemistry dataset). We find that forgetting is only weakly related to the preservation of the top-$k$ singular-value subspace, explaining why OP-LoRA performs similarly to the LoRA baseline.}
\label{tab:op_lora_rho}
\end{table}

Across all values of $k$, $\rho_k$ peaks at roughly $1.7\%$, meaning at least $98.3\%$ of the update's energy already lies in the orthogonal subspace. This indicates that forgetting is only weakly tied to the top-$k$ singular-value subspace of the pretrained weights.

\section{Implementation Details} \label{sec:impl_details}

Each experiment is averaged over three seeds. Hyperparameters are chosen after a sweep of learning rate $\in \{3\mathrm{e}{-6},\, 1\mathrm{e}{-5},\, 3\mathrm{e}{-5},\, 1\mathrm{e}{-4},\, 3\mathrm{e}{-4}\}$, batch size $\in \{8,\, 16,\, 32,\, 64\}$, and adapter rank $\in \{8,\, 32,\, 128,\, 512\}$, optimizing performance on the finetuning task.
Since the lora-learning phase is identical both in the GMM and LoRA baselines, we use the optimal LoRA parameters for the GMM runs as well. These are displayed in Table~\ref{tab:shared_hparams} and Table~\ref{tab:per_task_hparams}. For the GMM we sweep $K \in \{16,32\}$ and smoothing coefficient $\alpha \in \{0.1, 0.3, 1.0\}$

\begin{table}[h]
\centering
\caption{Shared hyperparameters across all tasks.}
\label{tab:shared_hparams}
\begin{tabular}{ll}
\toprule
\textbf{Hyperparameter} & \textbf{Value} \\
\midrule
\multicolumn{2}{l}{\textit{Optimizer (AdamW)}} \\
Weight decay & 0.1 \\
$\beta_1,\, \beta_2$ & 0.9,\, 0.95 \\
Gradient clip norm & 1.0 \\
\midrule
\multicolumn{2}{l}{\textit{Schedule}} \\
Warmup ratio & 0.1 \\
Decay ratio & 0.1 \\
Min learning rate & $\text{lr} / 100$ \\
\midrule
\multicolumn{2}{l}{\textit{GMM}} \\
JL projection dimension & 256 \\
GMM covariance type & Diagonal \\
\midrule
\multicolumn{2}{l}{\textit{Other}} \\
Sequence length & 1024 \\
Gradient checkpointing & Yes \\
LoRA $\alpha$ & $2r$ \\
\bottomrule
\end{tabular}
\end{table}

\begin{table}[h]
\centering
\caption{Per-task hyperparameters.}
\label{tab:per_task_hparams}
\begin{tabular}{lccc}
\toprule
& \textbf{Chemistry} & \textbf{Igbo Translation} & \textbf{Cybersecurity} \\
\midrule
\multicolumn{2}{l}{\textit{LoRA Adapter}} \\
Learning rate & $1\mathrm{e}{-4}$ & $1\mathrm{e}{-4}$ & $1\mathrm{e}{-4}$ \\
Batch size & 32 & 32 & 16 \\
Adapter rank $r$ & 128 & 128 & 512 \\
Epochs & 3 & 3 & 1 \\
\midrule
\multicolumn{2}{l}{\textit{GMM Gate}} \\
GMM positive components & 32 & 16 & 32 \\
GMM negative components & 32 & 32 & 32 \\
Smoothing $\alpha$ & 0.1 & 0.3 & 0.3 \\
\bottomrule
\end{tabular}
\end{table}

Our GMM implementation relies on the Pomegranate library~\citep{schreiber2018pomegranatefastflexibleprobabilistic}, while the rest of the code is written in PyTorch~\citep{paszke2019pytorchimperativestylehighperformance}. 

For OP-LoRA we also sweep the Top-$k$ value for $k \in \{16,128,512\}$ and select the best one.

LwF is implemented via a KL-divergence regularization term. At each training step, an additional forward pass acquires the base model's logits, which are compared to the logits produced with the adapter weights. We sweep the regularization coefficient over $\{0.1, 0.3, 1.0, 3.0, 10.0\}$. 

The MLP gate has two hidden layers of dimension 256 and 128. The MLP optimizes a typical binary cross entropy loss, where the model tries to classify whether the current token embedding belongs the current distribution or to the small generic pretraining dataset - similar to the GMM gate. In contrast to the GMM, a batch of MLP training data includes data from both labels (the GMM learns one distribution at a time). The gate's loss is added as an auxiliary loss to the network's main objective with a scaling factor. Similar to the GMM we perform a hyperparameter sweep including the learning rate of the MLP $\{1e-3, 1e-4, 1e-5\}$, the loss' scaling factor $\{10.0, 1.0, 0.1\}$, and the smoothing filter's coefficient $\{0.1, 0.3, 1.0\}$.

During multiphase learning the models use a learning rate of $1e-4$, batch size of 32, and adapter rank of 128. LSL used 16 components for $\Phi_{pos}$ and 32 components for $\Phi_{neg}$, as well as a smoothing coefficient of 0.1. Like in the single-task finetuning experiment we report the average performance over 3 seeds, igbo translation and chemistry are trained over 3 epochs each and cybersecurity over 1 epoch. Due to the long runtimes of this experiment we cap the amount of eval samples of each benchmark to 300.

All model checkpoints are taken from the Huggingface Model Hub\footnote{https://www.huggingface.co/models}:
\begin{itemize}
    \item \texttt{Qwen/Qwen2.5-7B-Instruct}
    \item \texttt{Qwen/Qwen2.5-3B-Instruct}
    \item \texttt{Qwen/Qwen2.5-1.5B-Instruct}
\end{itemize}
Our code is based on the official Huggingface implementation.

\section{Theoretical Motivation - Full Derivation}
\label{app:theory_full}

\subsection{Formulation}
\noindent \textbf{Setting.} Similar to the problem formulation in Sec.~\ref{sec:prob_inv}, we study a learning problem where tasks arrive sequentially, and analyze a single weight matrix. Earlier tasks had inputs from distributions $Q\in\mathcal Q$; during training an update is learned for the current task's input distribution $P$; during test time the system is evaluated on inputs from all of $\mathcal Q$ and $P$. As in Fig.~\ref{fig:method} (Left), a gate $g$ decides whether to apply the update. It is trained on samples from $P$ and possibly some small buffer $Q_0$; the earlier distributions $\mathcal Q$ are unknown at training time. We assume that the domain of the input features is bounded. This is a realistic assumption, since almost any weight matrix processes inputs that went through a layer normalization block.\footnote{Normalized features occupy a lower-dimensional set inside the box; volume is taken on that domain. The bounds do not depend on this choice, but uniform negatives must be drawn on the actual feature domain, not the ambient box, which they would almost surely miss.} We assume a task's features occupy a proper subset of the domain and that within that subset $P$'s density is bounded below by $c>0$; all densities are bounded above by $M$. Beyond this we assume nothing about $P$, and nothing about $\mathcal Q$ or $Q_0$. The same symbol denotes a distribution and its density.

\noindent \textbf{Target.}
Apply the update where the input looks like the data it was fitted on:
\[
A^*=\operatorname{supp}P,\qquad g^*=\mathbf 1[x\in A^*].
\]
where $\operatorname{supp}P$ is the support of P. This is the minimum-volume set carrying all of $P$'s mass - the $\alpha=1$ case of \citet{scott2006minimum}. $A^*$ is unique.\footnote{point-wise, every input either has density at least c or density zero, nothing in between, so there is no fringe to trade at the margin.}
In case of an overlap between the distributions $P$ and $Q$, the policy is that $P$ wins: a point in $A^*$ receives the update whichever distribution produced it, since later updates carry more recent information.

\noindent \textbf{Error.}
A gate is its accept set $\hat A$; its error on $Q$ is $Q(\hat A\triangle A^*)$, where $\hat A\triangle A^*=(\hat A\setminus A^*)\cup(A^*\setminus\hat A)$ is the symmetric difference - the set of points on which the gate and the target disagree. Since $\mathcal Q$ is unknown, the gate is judged against the worst admissible test distribution, and that worst case is exact:
\[
\sup_{Q\le M}\ \mathrm{err}_Q(\hat A)\;=\;\min\{1,\ M\,\mathrm{vol}(\hat A\triangle A^*)\},
\]
the supremum being attained by a $Q$ of density $M$ concentrated on the disagreement set. Minimizing the volume of disagreement with $A^*$ is therefore the minimax objective, not merely an upper bound.

\subsection{The gate's training objective is blind to the excess}

The symmetric difference has two parts. The \emph{deficit} is the part of $A^*$ the gate rejects. 
This is handled explicitly by the fitting procedure, and we do not discuss it further. The \emph{excess} is what the gate accepts outside $A^*$:
\[
E(\hat A)=\hat A\setminus A^*,\qquad \sup_{Q\le M}\ Q\big(E(\hat A)\big)=\min\{1,\ M\,\mathrm{vol}(E(\hat A))\},
\]
We are mainly interested in the excess because $P$ has no mass there, so any objective that is a $P$-weighted average of a per-point loss (e.g. a training loss) is blind to it \citep{steinwart2005classification}.

Intuition for a problematic scenario (not a proof): take any trained gate from a ``sufficiently expressive'' hypothesis class and change it to accept an arbitrary region outside $A^*$. It has the exact same $P$-averaged objective value (the changed region has no $P$-mass), yet a $Q$ concentrated on this region is now accepted in full. Since the objective cannot tell the two gates apart, which of them training returns is decided by something other than the data.

\subsection{Controlling what the objective cannot control} 
\label{app:theory_estimator_fams}
From the above, the training objective of the gate supplies no control over the excess. Whatever controls it therefore comes from elsewhere: the hypothesis class, the estimation procedure, or the optimizer and its regularization. Estimation families supply this control in different ways and at different costs; the following is not exhaustive.

\noindent \textbf{Local estimators.} Spheres of radius $r$ around the samples, or a kernel density thresholded. Every sample lies in $A^*$, so a union of spheres has excess confined to an $r$-shell around $A^*$ with no assumption on the shape of $P$ (containment in the shell, not a rate). The costs are memory growing with $n$ and coverage: gaps inside $A^*$ are rejected until samples fill them, which in high dimension requires many samples or a large $r$, widening the shell.

\noindent \textbf{Density estimation.}
Accept where an estimated density exceeds a threshold, $\hat A=\{\hat P\ge\tau\}$ with $0<\tau<c$\footnote{The true density jumps from $0$ outside $A^*$ to at least $c$ inside, so any $\tau$ in the gap gives the same set; it only sets the margin for estimation error on each side (a bump of $\hat P$ above $\tau$ outside is accepted, a dip below $\tau$ inside is rejected). Any fixed $\tau$ strictly inside $(0,c)$ is consistent.}.
Split the $L^1$ error over the excess $E=\hat A\setminus A^*$, the deficit $D=A^*\setminus\hat A$, and the rest:
\[
\|\hat P-P\|_1=\int_{E}|\hat P-P|+\int_{D}|\hat P-P|+\int_{\text{rest}}|\hat P-P|.
\]
On $E$ the gate accepts, so $\hat P\ge\tau$, while $P=0$: the integrand is at least $\tau$, giving $\int_E|\hat P-P|\ge\tau\,\mathrm{vol}(E)$. On $D$ the gate rejects, so $\hat P<\tau$, while $P\ge c$: the integrand is at least $c-\tau$, giving $\int_D|\hat P-P|\ge(c-\tau)\,\mathrm{vol}(D)$.  From the split, keeping $E$ and $D$ in the same inequality:
\[
\|\hat P-P\|_1\ \ge\ \tau\,\mathrm{vol}(E)+(c-\tau)\,\mathrm{vol}(D)\ \ge\ \min\{\tau,c-\tau\}\big(\mathrm{vol}(E)+\mathrm{vol}(D)\big),
\]
hence $\mathrm{vol}(\hat A\triangle A^*)\le\|\hat P-P\|_1/\min\{\tau,c-\tau\}$. The first inequality is the one that matters - the room left for an adversarial $Q$ is $M\,\mathrm{vol}(E)\le M\|\hat P-P\|_1/\tau$. An $L^1$-consistent density estimate therefore gives a consistent gate. \newline
In contrast to the union of spheres estimator, a fixed-size parametric family (e.g.\ a Gaussian mixture) has constant memory. However, this has a price - consistent density estimation requires $\|\hat P-P\|_1\to0$, which the family must allow; when it cannot, the excess can persist with unlimited data (misspecified ``two-mode" example below).

\textbf{Density with an input-dependent threshold.} The same argument holds for $\hat A = \{\hat P(x) \geq \tau(x)\}$ with $\tau_{\min} = \inf_x \tau(x)$ and $\tau_{\max} = \sup_{x \in A^*} \tau(x)$: the integrand is at least $\tau_{\min}$ on $E$ and at least $c - \tau_{\max}$ on $D$, so $\mathrm{vol}(E) \leq \|\hat P - P\|_1/\tau_{\min}$ whenever $\tau_{\min} > 0$, and $\mathrm{vol}(D) \leq \|\hat P - P\|_1/(c - \tau_{\max})$ if also $\tau_{\max} < c$ (a constant $\tau$ recovers the bound above).
LSL is this gate with $\hat P = \Phi_{\mathrm{pos}}$ and $\tau = \Phi_{\mathrm{neg}}$. Choosing $\tau(x)$ so that the bound is tight is not trivial; $\Phi_{\mathrm{neg}}$ is a practical approximation that satisfies $\tau_{\min} > 0$ by construction, since a Gaussian mixture is positive on the bounded domain. Empirically, the resulting gate opens on about 16\% of tokens from out-of-distribution pretraining tasks (5\% with temporal smoothing) while retaining 96.6\% of pretrained performance (98.8\% with smoothing; Fig.~\ref{fig:gmm_ablation_ncomps_temp_smooth}).

\noindent \textbf{Classifier.} Train a classifier to separate $P$ (positives) from a buffer $Q_0$ (negatives). Under equal priors its population accept set is $\{P\ge Q_0\}$ - wherever $Q_0$ has mass it rejects unless $P$ is denser, and wherever neither has mass the objective is indifferent and the decision is left to other factors. Which set this is depends entirely on $Q_0$.

\begin{itemize}
\item \emph{Generic buffer} (some data). Outside $A^*$ it rejects where the buffer sits and is unconstrained everywhere else: that region is exposed. Inside $A^*$, the classifier withholds the update wherever $Q_0>P$, against the policy. This is the class of the MLP gate tested in Sec.~\ref{sec:ablations}, which we found empirically to be suboptimal.

\item \emph{Buffer drawn from earlier tasks.} The exposed region shrinks as the buffer covers more of $\mathcal Q$, and vanishes when the buffer covers every earlier distribution outside $A^*$. This requires both access to $\mathcal Q$ and being able to cover $\mathcal Q$ at each training phase - two assumptions that are impractical for large-scale pretraining settings.

\item \emph{Uniform over the domain.} assuming $Q_0\equiv u$ with $u=1/\mathrm{vol}(\text{domain})<c$: the accept set is $\{P\ge u\}=A^*$ exactly. This is minimum-volume estimation - reject uniform points subject to covering $P$ - the target's own definition \citep[\S8]{scott2006minimum} and the learner of \citet[Thm.~9]{fang2022ood}. It needs a hypothesis class that approximates $A^*$ with controlled complexity, and in high dimension few uniform points land near a small $A^*$, requiring a very large sample. In effect it is support estimation by an expensive route.

\end{itemize}

Classification against an arbitrary reference distribution therefore does not, by itself, guarantee support recovery.
The input-dependent-threshold bound does not apply to a general classifier, whose score need not estimate $P$ and whose threshold need not be bounded away from zero.

\noindent \textbf{Examples.} Below are three one-dimensional cases that make the failure modes concrete: a parametric family that cannot represent $P$ and so accepts a gap forever; two densities that likelihood cannot tell apart but whose gates differ; and a classifier with zero training error that accepts a region of volume ten outside the support.

\begin{itemize}
\item \emph{Misspecified.} $P=\tfrac12\mathrm{Unif}[-3,-2]+\tfrac12\mathrm{Unif}[2,3]$; a single Gaussian fitted by maximum likelihood converges to $\mathcal N(0,19/3)$, and any level set covering both modes covers the gap. For $Q=\mathrm{Unif}[-1,1]$ the excess is $1$ with unlimited data.
\item \emph{Unidentified.} $P=\mathrm{Unif}[0,1]$, threshold $0.05$, and a class of two densities $f_0=0.9\cdot\mathbf 1_{[0,1]}+0.025\cdot\mathbf 1_{[2,6]}$, $f_1=0.9\cdot\mathbf 1_{[0,1]}+0.1\cdot\mathbf 1_{[7,8]}$. They have identical likelihood on every sample from $P$, yet $\{f_0\ge0.05\}=[0,1]$ and $\{f_1\ge0.05\}=[0,1]\cup[7,8]$: likelihood fixes how much mass sits off the data, not where. (The class is misspecified - the true density beats both - so this concerns flexible off-data mass, not maximum likelihood in a correct family.)
\item \emph{Classifier.} $P=\mathrm{Unif}[0,1]$ against a buffer $Q_0=\mathrm{Unif}[2,3]$ on the domain $[-10,10]$. Every threshold in $(1,2)$ separates the two perfectly, so the classifier is unidentified among them; and every one of them accepts $[-10,0]$, outside $\operatorname{supp}P$ - excess of volume $10$ from a classifier with zero training error.
\end{itemize}

\noindent \textbf{Conclusion.}
While every training objective on $P$ is blind to the excess, we show that the disagreement volume can be controlled by something one can actually minimize: $L^1$ density error, or coverage plus volume. Every family that controls the excess supplies that missing quantity from outside the data, and pays for it. A parametric density supplies it through its family, at constant memory, and pays in assumptions: when the family cannot represent $P$, the excess can persist. A local estimator supplies it through its radius with no assumption on $P$, and pays in memory that grows with the data. A uniform buffer supplies it through a volume count, and pays in reference samples, potentially many in high dimension.

\section{Fitting the GMM and the Weight Adapter}
\label{sec:gmm_fitting_algo}

While fitting a single GMM is conceptually simple (standard EM), fitting more than one hundred GMMs (one per adapter) in high dimensions over a large dataset is computationally demanding. Furthermore, unlike gradient descent, conventional EM requires access to the full dataset at each step. To alleviate this, we fit GMMs module-by-module, as described in Algorithm~\ref{alg:fit_gmm}: since each GMM depends only on its own module's inputs, this computation is exact and allows us to hold inputs for only one module at a time, discarding them before proceeding to the next.

\begin{algorithm}[H]
\caption{FitGMMs} \label{alg:fit_gmm}
\begin{algorithmic}[1]
\Require Model $M$ with $L$ layers and $J$ modules per layer, dataset $\mathcal{D}$
\Ensure Trained GMMs $\boldsymbol{\Phi} = \{\Phi^{(l,j)}\}_{l \in [L], j \in [J]}$
\Statex
\State Load full dataset $\mathcal{D}$
\For{$l = 1, \ldots, L$}
    \For{$j = 1, \ldots, J$}
        \State Compute all inputs to module-$(l,j)$; discard previous inputs
        \State Fit $\Phi^{(l,j)}$ on inputs via EM
    \EndFor
\EndFor
\State \Return $\boldsymbol{\Phi}$
\end{algorithmic}
\end{algorithm}

Fitting the weight adapter is identical to standard adapter finetuning. The key insight is that during training, all data is from the new task's distribution, so the gate should remain open throughout the training phase. This design avoids the need for bootstrapping - without it, $\Phi_{pos}$ would need to be initialized before the adapter has learned the new representation. We note that this introduces a train-inference mismatch (the gate may not have perfect accuracy on the new phase's data), but we find empirically that this does not affect performance.

\begin{algorithm}[H]
\caption{FitAdapters} \label{alg:fit_adapter}
\begin{algorithmic}[1]
\Require Model $M$, fine-tuning data $\mathcal{D}_{i}$, number of epochs $E$
\Ensure Adapter weights $\mathbf{W}_{\text{adapter}}$
\Statex
\State Apply new adapter weights $\mathbf{W}_{\text{adapter}}$ to $M$ 
\For{$e = 1, \ldots, E$}
    \For{each batch $B \in \mathcal{D}_{i}$}
        \State Forward pass through $M$ \Comment{Gate always open during training}
        \State Compute loss $\mathcal{L}$
        \State Update $\mathbf{W}_{\text{adapter}}$ via backpropagation
    \EndFor
\EndFor
\State \Return $\mathbf{W}_{\text{adapter}}$
\end{algorithmic}
\end{algorithm}

\section{Why per-matrix gates? - Extended Discussion}
\label{app:single_router}

In Sec.~\ref{sec:ablations} and Fig.~\ref{fig:single_router_ablation} we showed that a single per-token gate after the embedding layer, controlling all adapters, falls short of LSL's per-matrix gates in both learning and retention. Here we discuss further alternatives. A single gate at an intermediate layer would see contextual features, but adapters below that layer need its decision before it is computed, requiring an additional forward pass through the preceding layers; per-matrix gates obtain contextual decisions in a single pass. Finally, we avoid per-sequence gating: a sequence, and more generally a stream, can mix data from several domains without given boundaries, and only per-token decisions can follow such changes.

\newpage
\section{Function Behavior Change} \label{sec:fbc}
\begin{figure}[h]
  \centering
  \includegraphics[width=\textwidth]{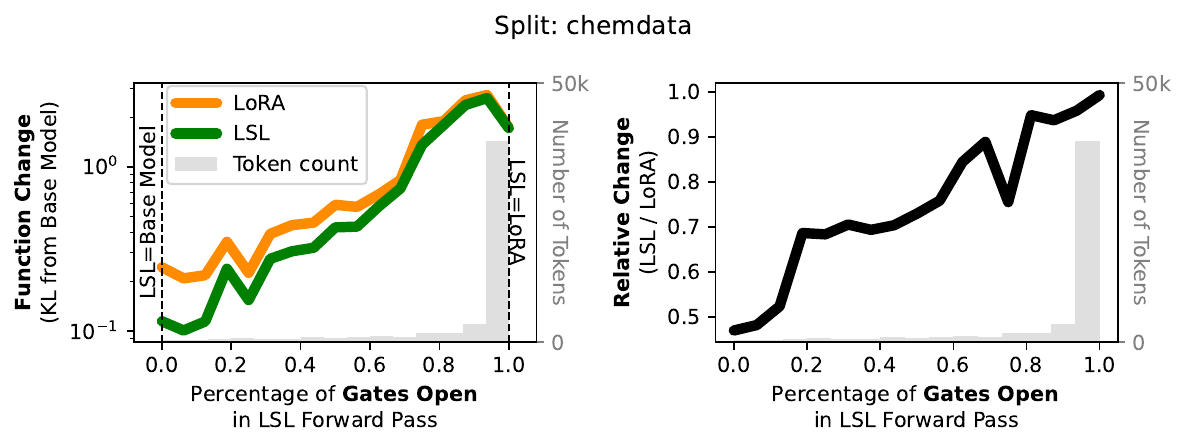}
  \caption{\textbf{Function Behavior Change - In-Distribution Data.} For each token of the finetuning (chemistry) data, we measure the KL divergence between the outputs of the finetuned and base models, as a function of the fraction of LSL gates open on that token. Most tokens open nearly all gates, where LSL and LoRA change the model's behavior similarly, as desired.}
  \label{fig:fbc_id}
  \vspace{-14pt}
\end{figure}

\begin{figure}[h]
  \centering

  \includegraphics[width=\textwidth]{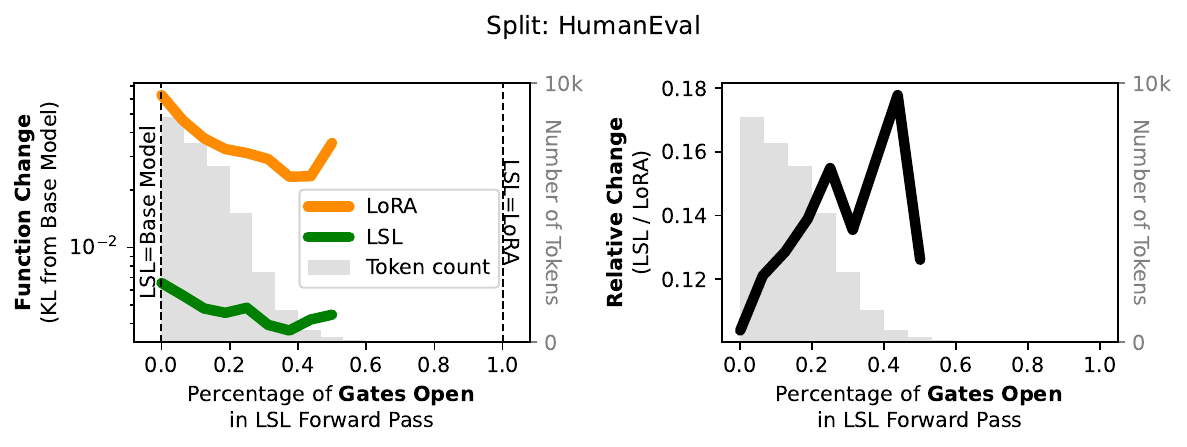}
  \includegraphics[width=\textwidth]{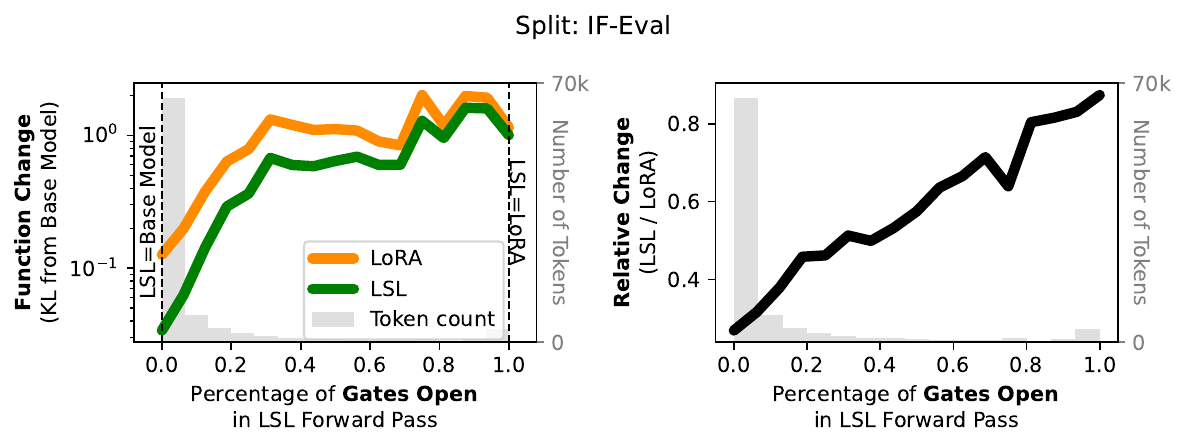}
  \includegraphics[width=\textwidth]{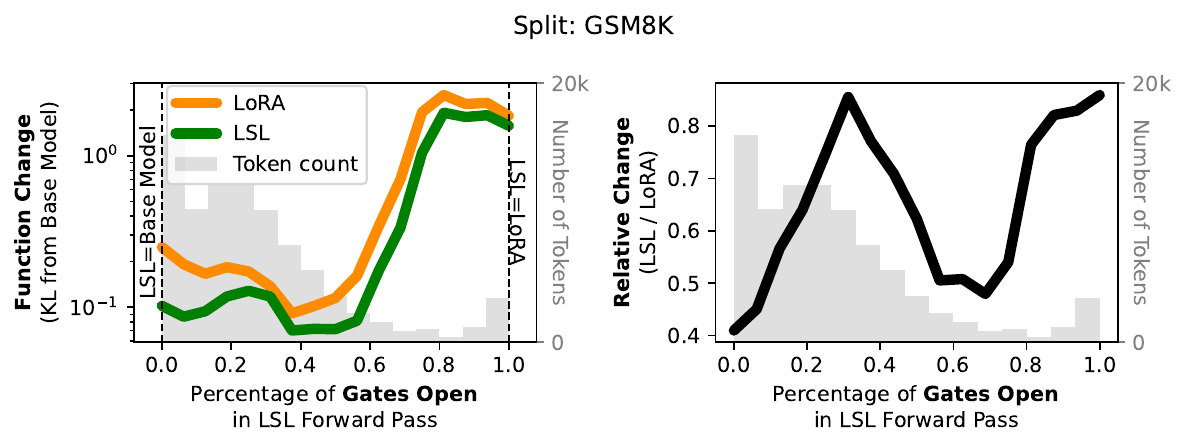}
  \caption{\textbf{Function Behavior Change - Out-of-Distribution Data.} Same analysis on pretraining benchmarks. Here most tokens open few gates, and LSL changes the model's behavior less than LoRA, sometimes by an order of magnitude. The LoRA baseline changes the model's behavior more on these out-of-distribution tokens, degrading pretraining capabilities.}
  \label{fig:fbc_ood}
  \vspace{-14pt}
\end{figure}

To measure the gate's impact directly in function space, we compute, for each token, the KL divergence between the output distributions of the finetuned and base models, and group tokens by the fraction of LSL gates open on them in the forward pass (model finetuned on chemistry). On in-distribution data (Fig.~\ref{fig:fbc_id}), most tokens open nearly all gates, and LSL and LoRA change the model's behavior similarly, as intended. On pretraining benchmarks (Fig.~\ref{fig:fbc_ood}), most tokens open few gates, and LSL changes the model's behavior far less than LoRA - whereas LoRA's changes on these inputs degrade pretraining capabilities.

\newpage
\section{Union of Spheres - Radius Calibration} \label{sec:uos_r_calib}
\begin{figure}[h]
  \centering
  \includegraphics[width=\textwidth, trim = 0 0 0 40, clip]{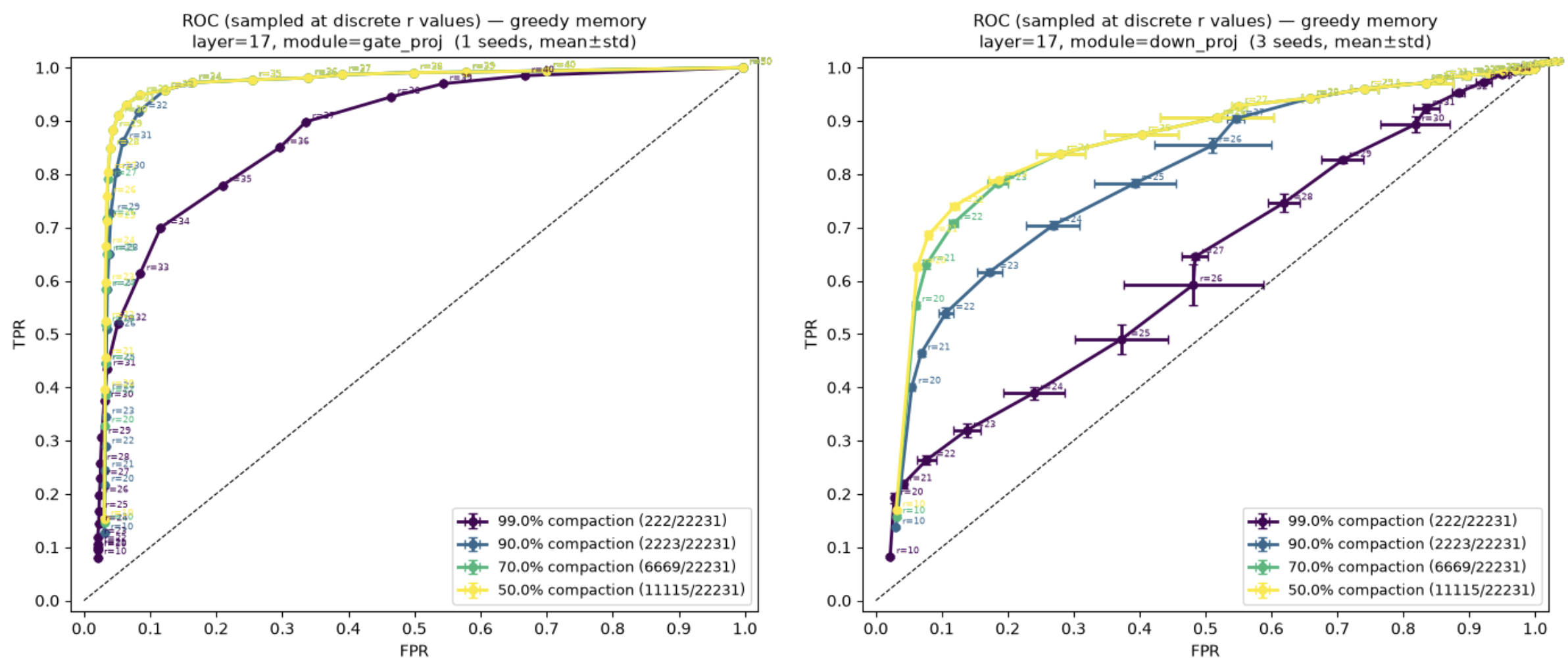}
  \caption{\textbf{ROC Curves for UoS Radius Calibration}. We show two representative curves, one for the gate projection (left) and one for the down projection (right), both in layer 17. Each curve color indicates different compaction rates (dictated by the maximal value of representative points M). The selected radius is the one with the maximal compaction rate, subject to being within the allowed TPR,FPR region (defined globally for all gates).}
  \label{fig:uos_r_calib_roc_curves}
\end{figure}

The support $\mathcal{M}$ is defined using a predefined radius $r$ with respect to the representative subset. We find that a global $r$ for all UoS gates does not perform well due to the different dimensions (e.g. the dimension of the down projection matrix of the MLP is about 4 times larger than the dimension of the other matrices), and different input distributions between layers. To find the optimal $r$ per gate we propose a one-time calibration process, detailed below.
The algorithm relies on two sets of points: A small set from the current phase's data, and a small random sample from the generic pre-training data. We first record the activations of the pre-trained model for both sets. Then, per gate, we compute a ROC curve for different values of r (see example in Fig.~\ref{fig:uos_r_calib_roc_curves}).
Notice that there are several curves per gate - each one is for a different compaction ratio - namely, different values for the upper limit of number of representatives $M$. 
The optimal r is the one that gives the maximal compaction rate under the constraint that it is above a global TPR threshold and below a global FPR threshold. For our experiment we found that $\text{TPR}_{\text{TH}}=0.17,\text{FPR}_{\text{TH}}=0.03$ gives optimal results.

\begin{figure}[h]
  \centering
  \includegraphics[width=\textwidth, trim=55 475 55 50, clip]{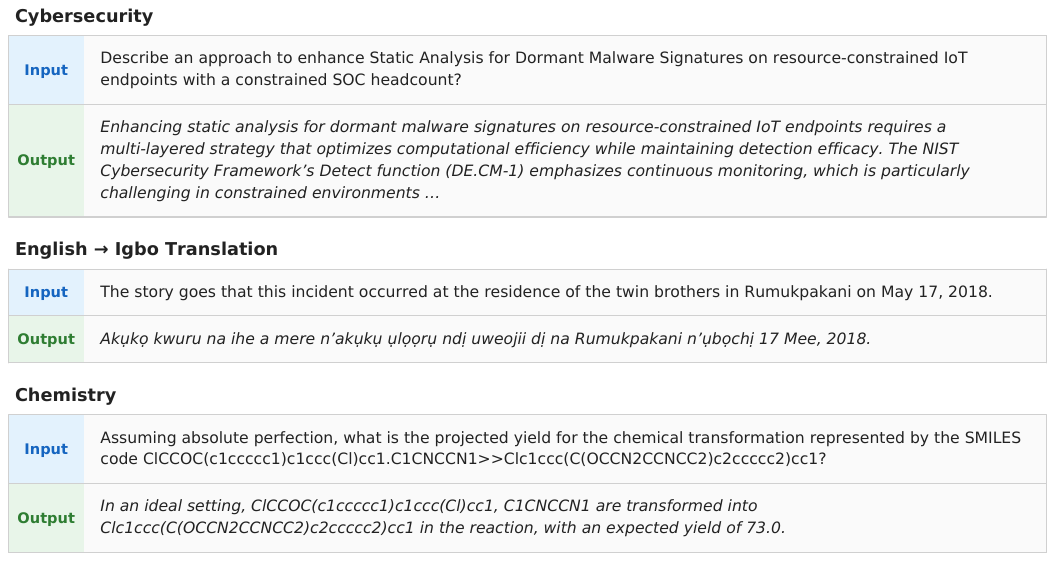}
  \caption{\textbf{Samples from the finetuning datasets}.}
  \label{fig:dataset_samples}
\end{figure}

\end{document}